\documentclass[10pt,twocolumn,letterpaper]{article}
\usepackage[pagenumbers]{cvpr}  % To force page numbers, e.g. for an arXiv version
\usepackage[accsupp]{axessibility}

\newcommand{\E}{\mathbb{E}}
\newcommand{\Var}{\mathrm{Var}}
\newcommand{\N}{\mathcal{N}}
\newcommand{\R}{\mathbb{R}}

\definecolor{cvprblue}{rgb}{0.21,0.49,0.74}
\usepackage[pagebackref,breaklinks,colorlinks,allcolors=cvprblue]{hyperref}

\usepackage{multirow}
\usepackage{booktabs}
\usepackage{array}
\usepackage{float}
\usepackage{diagbox}
\newcommand{\paragraphtitle}[1]{\noindent \textbf{#1}}
\usepackage{amsmath}
\usepackage[linesnumbered,ruled,vlined]{algorithm2e}
\usepackage{graphicx}
\usepackage{amssymb}
\usepackage{soul}
\usepackage[table]{xcolor}
\usepackage{mathtools}
\definecolor{tabfirst}{rgb}{1, 0.7, 0.7}
\definecolor{tabsecond}{rgb}{1, 0.85, 0.7}
\definecolor{tabthird}{rgb}{1, 1, 0.7}
\newcommand{\first}[1]{\cellcolor{tabfirst}#1}
\newcommand{\second}[1]{\cellcolor{tabsecond}#1}
\newcommand{\third}[1]{\cellcolor{tabthird}#1}
\newcommand{\highlightinline}[2]{\begingroup\setlength{\fboxsep}{1pt}\colorbox{#1}{#2}\endgroup}
\usepackage{amsthm}
\newtheorem{assumption}{Assumption}
\newcommand{\indep}{\perp\!\!\!\perp}
\usepackage{makecell}
\usepackage{booktabs}
\usepackage{rotating}

\def\confName{CVPR}
\def\confYear{2026}
\title{Evidential Neural Radiance Fields}
\author{
    Ruxiao Duan\\
    Yale University\\
    {\tt\small ruxiao.duan@yale.edu}
    \and
    Alex Wong\\
    Yale University\\
    {\tt\small alex.wong@yale.edu}
}

\begin{document}
\maketitle

\begin{abstract}
Understanding sources of uncertainty is fundamental to trustworthy three-dimensional scene modeling. While recent advances in neural radiance fields (NeRFs) achieve impressive accuracy in scene reconstruction and novel view synthesis, the lack of uncertainty estimation significantly limits their deployment in safety-critical settings. Existing uncertainty quantification methods for NeRFs fail to separately capture both aleatoric and epistemic uncertainties. Among those that do quantify one or the other, many of them either compromise rendering quality or incur significant computational overhead to obtain uncertainty estimates. To address these issues, we introduce Evidential Neural Radiance Fields, a probabilistic approach that seamlessly integrates with the NeRF rendering process, enabling direct quantification of both aleatoric and epistemic uncertainties from a single forward pass. We compare multiple uncertainty quantification methods on three standardized benchmarks, where our approach demonstrates state-of-the-art scene reconstruction fidelity and uncertainty estimation quality. Code is available at https://github.com/KerryDRX/EvidentialNeRF.
\end{abstract}

\begin{figure*}[t]
    \centering
    \includegraphics[width=\linewidth]{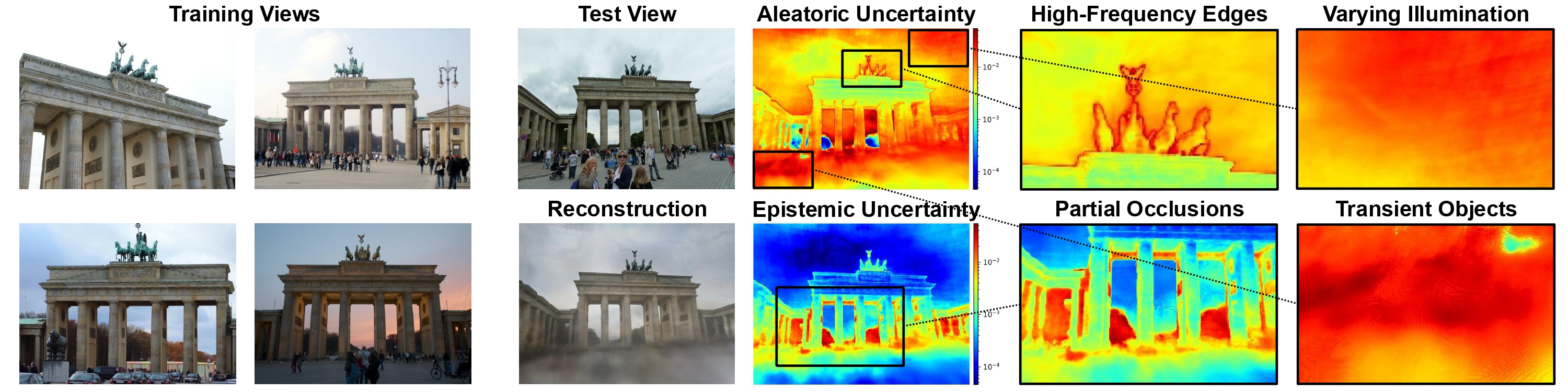}
    \vspace{-7mm}
    \caption{
        Scene reconstruction along with its aleatoric and epistemic uncertainties rendered by Evidential NeRF from a scene in the wild.
        AU arises from intrinsic data variability in the training images, including radiance variation (e.g., sky regions due to changing illumination), high-frequency regions (e.g., object edges), and the presence of transient objects (e.g., pedestrians).
        EU indicates the model's lack of knowledge, prominently appearing in occluded regions where training supervision is insufficient (e.g., the trees occluded by the gate).
    }
    \vspace{-4mm}
    \label{fig:teaser}
\end{figure*}

\section{Introduction}

Predictive uncertainty of deep neural networks originates from two different sources, data and model, and the two types of uncertainty are respectively referred to as aleatoric uncertainty (AU) and epistemic uncertainty (EU)~\cite{normal,aeuintroduction}.
Aleatoric (data) uncertainty stems from intrinsic randomness of the data generation process, while epistemic (model) uncertainty arises from model's lack of knowledge~\cite{uqreview2,uqreview3}.
Understanding both types of predictive uncertainty is essential for building models that are not only accurate but also reliable and explainable under various conditions.

Neural radiance fields (NeRFs)~\cite{nerf} have achieved remarkable performance in three-dimensional (3D) scene reconstruction and novel view synthesis, yet their incapability of quantifying predictive uncertainty poses significant challenges for their broader adoption in safety-critical domains such as autonomous driving~\cite{nerf_driving}, medical imaging~\cite{nerf_medical}, and robotics~\cite{nerf_robotics}, where precise and prompt quantification of predictive uncertainty remains crucial.

We consider an essential but overlooked research question: \textit{How should a NeRF report what it does not know (epistemic uncertainty) versus what the data cannot resolve (aleatoric uncertainty), without sacrificing rendering fidelity or speed?}
Existing uncertainty quantification (UQ) methods for NeRFs generally fall into three categories: closed-form likelihood models, which are unable to capture epistemic uncertainty~\cite{nerfw,activenerf,mixnerf,flipnerf}, Bayesian methods, which typically necessitate sampling during inference~\cite{mcdropout,snerf,cfnerf}, and ensemble approaches, which require training multiple models~\cite{ensembles,dane}.
None of these methods quantifies both types of uncertainty with a single forward pass, and many of them have to sacrifice prediction accuracy to obtain uncertainty estimates.
Among the likelihood models is the classical Gaussian approach~\cite{normal} that represents a prediction as the mean and uncertainty as the variance of a normal distribution.
This method has been widely adopted in not only NeRFs~\cite{nerfw,activenerf,nerfonthego} but also other radiance field frameworks such as Gaussian Splatting~\cite{3dgs,wildgaussians} and DVGO~\cite{dvgo,uncertaintyfield}.
While computationally efficient, this paradigm inherently models only aleatoric uncertainty while failing to account for epistemic uncertainty.

To address these limitations, we propose Evidential Neural Radiance Fields (Evidential NeRFs), a probabilistic framework to separately quantify both aleatoric and epistemic uncertainties of NeRFs through a single forward pass (\Cref{fig:teaser}).
Our method extends the normal probabilistic formulation of radiance modeling by treating the predicted mean and variance of pixel radiance themselves as random variables governed by a higher-order evidential distribution, yielding closed-form predictive uncertainties through the NeRF rendering process.
Unlike prior evidential deep learning methods~\cite{edl,der}, which regress evidential distribution parameters, we adapt evidential reasoning to the volumetric structure of NeRFs, enabling them to predict aleatoric and epistemic uncertainties directly instead of having to reformulate them from evidential parameters. 

Beyond the technical limitations of existing approaches, which we address with Evidential NeRF, we also observed that current uncertainty quantification benchmarks lack standardization.
Methods are often evaluated using different architectures, data splits, and training setups, making direct comparisons difficult.
To mitigate these confounding factors, we establish a new benchmark to isolate and evaluate the UQ methods themselves.
Under this standardized evaluation, our method consistently ranks within the top three across all image reconstruction and uncertainty quantification metrics, if not the first.
The remaining top-performing methods are mostly ensemble-based, which incur substantial computational cost and are unsuitable for real-time applications.
In contrast, our approach is the second-fastest overall, being only 0.04 FPS slower than the fastest method while delivering significantly better performance across all metrics.

Our contributions can be summarized as follows.
\begin{itemize}
    \item We present a probabilistic framework for neural radiance fields to separately quantify both aleatoric and epistemic uncertainties in 3D scene reconstructions.
    \item We provide detailed mathematical derivations showing how aleatoric and epistemic uncertainties can be propagated from points to pixels under proper independence assumptions, enabling seamless integration of evidential deep learning with the volumetric rendering paradigm.
    \item We establish a standardized benchmark for faithful comparison of NeRF uncertainty quantification methods.
\end{itemize}

\begin{figure*}[t]
    \centering
    \includegraphics[width=\linewidth]{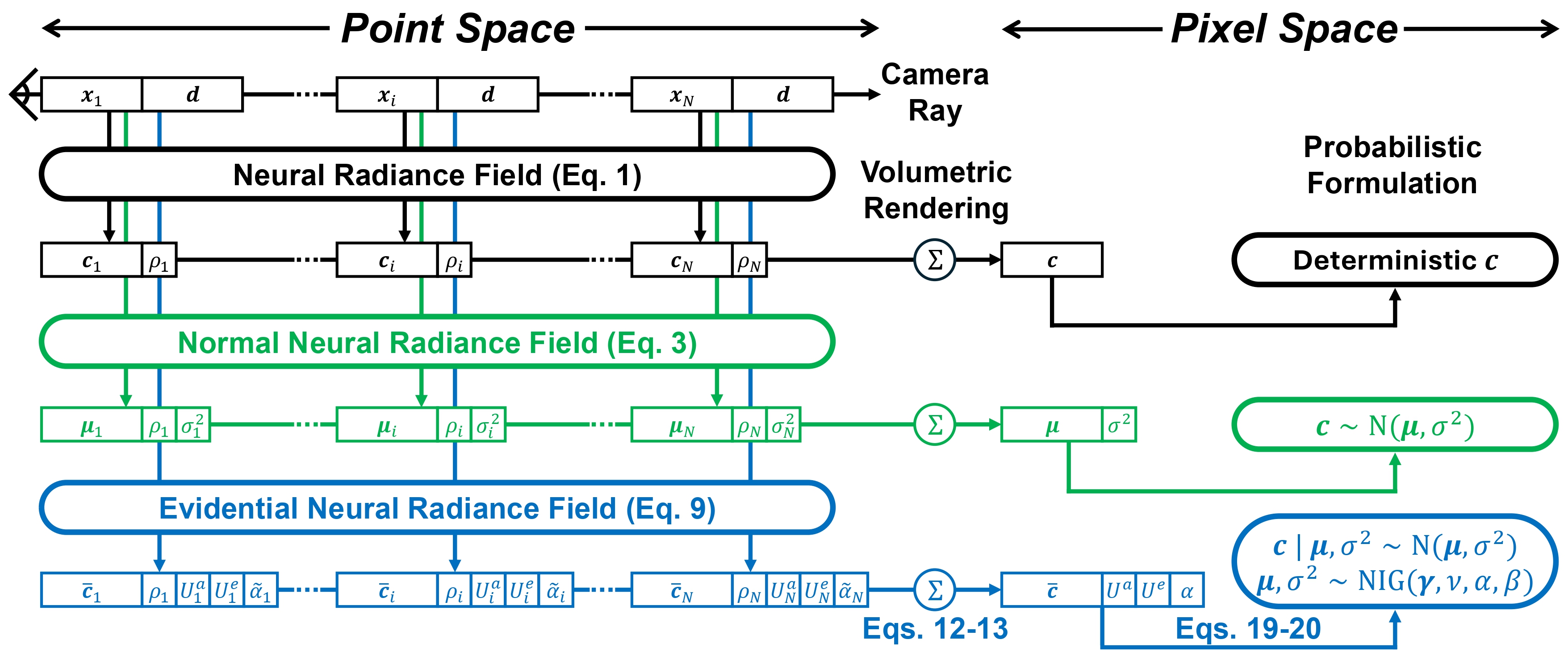}
    \caption{
        Evolution of NeRF pipeline across three levels of probabilistic formulations.
        $N$ points sampled along the camera ray give $N$ pairs of spatial location and viewing direction, which are passed to the NeRF model for prediction.
        Level 1: Vanilla NeRF predicts only point density and color, resulting in a deterministic pixel color without any uncertainty estimate.
        Level 2: Normal NeRF assumes the point and pixel colors follow normal distributions, quantifying aleatoric uncertainty of rendered color.
        Level 3: Evidential NeRF assumes the pixel color has random mean and variance following an evidential distribution, quantifying both aleatoric and epistemic uncertainties.
    }
    \vspace{-0.4cm}
    \label{fig:method}
\end{figure*}

\section{Related Work}
\label{sec:related_work}

\paragraphtitle{Neural radiance fields.}
NeRFs~\cite{nerf} represent a 3D scene as a plenoptic function which can be learned from a sparse set of input views.
Realized as a deep neural network, the plenoptic function maps spatial locations and viewing directions to their volume densities and radiance.
NeRFs produce high-fidelity reconstructions and photorealistic novel view renderings. Numerous variants in architectures, loss functions, and sampling strategies have been proposed to further improve the model's training efficiency, inference speed, and rendering quality~\cite{regnerf,dietnerf,mobilenerf,tensorf,refnerf,instantngp,mipnerf,mipnerf360,fastnerf,efficientnerf,pointnerf,zipnerf}.

\paragraphtitle{NeRF uncertainty quantification.}
UQ methods for NeRFs generally follow three schools of thought.

First, closed-form likelihood models.
NeRF-W~\cite{nerfw} and ActiveNeRF~\cite{activenerf} follow the same probabilistic framework in~\cite{normal} to model ray colors by normal distributions.
MixNeRF~\cite{mixnerf} and FlipNeRF~\cite{flipnerf}, though not proposed as UQ methods, model ray colors by mixtures of Laplace distributions to represent the multimodal nature of radiance.
These approaches employ closed-form probabilistic distributions to model pixel colors, enabling efficient training and uncertainty inference, but overlooking epistemic uncertainty, due to the deterministic nature of the distributions employed.

Second, Bayesian methods.
Monte Carlo dropout~\cite{mcdropout} estimates epistemic uncertainty by interpreting dropout as variational inference and performing multiple stochastic forward passes at test time.
S-NeRF~\cite{snerf} and CF-NeRF~\cite{cfnerf} learn a probability distribution over all the possible radiance fields and approximate the posterior distribution with stochastic variational inference~\cite{vi} and conditional normalizing flows~\cite{normalizingflows}.
While these approaches provide epistemic uncertainty estimates, they typically incur significant computational overhead due to sampling at inference.

Third, ensemble methods.
Deep ensembles~\cite{ensembles} estimate epistemic uncertainty by training multiple models independently and measuring their prediction variance during inference.
Density-aware NeRF Ensembles (DANE)~\cite{dane} further incorporates a density-based epistemic term that captures uncertainty in scene geometry and appearance, which is added to the uncertainty estimated by standard ensembles.
Ensemble-based methods impose the most significant computation and memory overheads, as each model must be trained, stored, and separately evaluated during inference.

\paragraphtitle{Evidential deep learning.}
Based on the theory of subjective logic~\cite{subjectivelogic}, evidential deep learning (EDL)~\cite{edl} is first introduced for classification uncertainty quantification based on a single network and later extended to regression~\cite{der}.
EDL considers model learning as a process of evidence collection, in which each training sample adds support to a higher-order evidential distribution, which is a Dirichlet distribution for classification and a normal-inverse-gamma (NIG) distribution for regression~\cite{edl,der}.
The probability distribution from which predictions are drawn is assumed to have random parameters sampled from the evidential distribution, and the aleatoric and epistemic predictive uncertainties can be directly obtained by a single forward propagation.
Though effective in standard regression, EDL is not naively compatible with NeRF learning paradigm due to its hierarchical volumetric formulation: As NeRFs receive supervision after volumetric rendering, evidential parameters tied to point-level predictions cannot be learned directly from pixel-level observations.
Concurrent with our work, ENeRF~\cite{enerf} proposes evidential modeling in NeRFs by adopting NIG distribution at the point level and approximating pixel-level parameters via the mixture of NIGs~\cite{monig}.

\section{Preliminaries}
\label{sec:preliminaries}

\paragraphtitle{Neural radiance fields.}
NeRF learns a continuous scene representation function $f$ by a multilayer perceptron that maps a pair of spatial location (referred to as point in this paper) $\boldsymbol{x}_i \in \R^3$ and unit viewing direction $\boldsymbol{d} \in \R^3$ to its corresponding view-independent volume density $\rho_i > 0$ and directional emitted radiance $\boldsymbol{c}_i \in [0, 1]^3$ in RGB color.
For notational simplicity, consider a scalar color channel $c_i \in [0, 1]$.
A NeRF model predicts
\begin{equation}
\label{eq:nerf}
    c_i, \rho_i = f(\boldsymbol{x}_i, \boldsymbol{d}).
\end{equation}
The pixel color of a ray $\boldsymbol{r}(t) = \boldsymbol{o} + t \boldsymbol{d}$ defined by a camera center $\boldsymbol{o}$ and a viewing direction $\boldsymbol{d}$ is derived by first sampling $N$ points $\{ \boldsymbol{r}(t_i) \}_{i=1}^N$ along the ray and then computing the weighted sum of point colors via the discretized volumetric rendering equation $c = \sum_{i=1}^N w_i c_i$, where $w_i = \exp\left( -\sum_{j=1}^{i-1}\rho_j\delta_j \right) \left(1-\exp(-\rho_i\delta_i) \right)$ is the weight of the $i$-th point and $\delta_i = t_{i+1} - t_i$ is the distance between two adjacent points.

\paragraphtitle{Normal neural radiance fields.}
Based on the probabilistic framework in~\cite{normal}, a point color $c_i$ can be modeled by a normal distribution as
\begin{equation}
    \label{eq:point_normal}
    c_i \ | \ \mu_i, \sigma_i^2 \sim \N(\mu_i, \sigma_i^2),
\end{equation}
with mean $\mu_i \in [0, 1]$ and variance $\sigma_i^2 > 0$ predicted by
\begin{equation}
    \label{eq:point_normal_function}
    (\mu_i, \sigma_i^2), \rho_i = f(\boldsymbol{x}_i, \boldsymbol{d}).
\end{equation}
As the pixel color $c$ is the weighted sum of point colors $c_i$'s, $c$ also follows a normal distribution assuming independence of points:
\begin{equation}
    \label{eq:pixel_normal}
    c \ | \ \mu, \sigma^2 \sim \N(\mu, \sigma^2),
\end{equation}
where $\mu \coloneq \sum_{i=1}^N w_i \mu_i$ is the mean (prediction) and $\sigma^2 \coloneq \sum_{i=1}^N w_i^2 \sigma_i^2$ is the variance (uncertainty) of pixel color.
While adjacent points in practice exhibit statistical dependence in radiance, the independence assumption is a common and effective simplification to enable tractable aggregation of point uncertainties into the pixel level.
Though efficient, this approach only captures aleatoric uncertainty.

\section{Method}

Based on~\Cref{eq:point_normal,eq:pixel_normal}, we establish an evidential probabilistic framework for radiance modeling by taking a step further to assume random conditional mean and variance.
Different from conventional EDL approaches which train the model to predict evidential parameters, we instead let the model predict aleatoric and epistemic uncertainties directly (\Cref{sec:method_point}), and then propagate the uncertainties from points to pixels compatible with the volumetric rendering paradigm of NeRFs (\Cref{sec:method_propagation}).
Then the evidential distribution can be reformulated from the rendered color and uncertainties (\Cref{sec:method_pixel}) and learned at the pixel level where supervision is available (\Cref{sec:method_learning}).

\subsection{Point-level Probabilistic Radiance Modeling}
\label{sec:method_point}

Building upon~\Cref{eq:point_normal}, we further treat the conditional mean and variance of point radiance as random variables instead of point estimates, i.e., $(\mu_i, \sigma_i^2) \sim \pi_i$ where $\pi_i$ is some probability distribution.
Under this formulation, the predictive mean, total, aleatoric, and epistemic uncertainties of the point color $c_i$ can be respectively formulated as
\begingroup
\allowdisplaybreaks
\begin{align}
    \bar{c}_i
    &\coloneq \E[c_i]
    = \E[\E[c_i | \mu_i, \sigma_i^2]]
    = \E[\mu_i],
    \\
    U_i
    &\coloneq \Var[c_i]
    = U_i^\text{alea} + U_i^\text{epis},
    \\
    U_i^\text{alea}
    &\coloneq \E[\Var[c_i | \mu_i, \sigma_i^2]]
    = \E[\sigma_i^2],
    \\
    U_i^\text{epis}
    &\coloneq \Var[\E[c_i | \mu_i, \sigma_i^2]]
    = \Var[\mu_i].
\end{align}
\endgroup

\paragraphtitle{Note:}
The classical assumption that $\mu_i$ and $\sigma_i^2$ are point estimates corresponds to the special case where $\pi_i$ is a degenerate distribution with a Dirac delta density function concentrated at fixed values of $\hat\mu_i$ and $\hat\sigma_i^2$ and the uncertainties above are thereby reduced to $U_i^\text{alea} = \E[\sigma_i^2] = \hat\sigma_i^2$ and $U_i^\text{epis} = \Var[\mu_i] = 0$, in which case only aleatoric uncertainty can be captured.

An Evidential NeRF model predicts
\begin{equation}
    \label{eq:f_evidential}
    (\bar{c}_i, U_i^\text{alea}, U_i^\text{epis}, \tilde{\alpha}_i), \rho_i = f(\boldsymbol{x}_i, \boldsymbol{d}),
\end{equation}
where $\bar{c}_i \in [0, 1]$ is the point's mean color, $U_i^\text{alea} > 0$ and $U_i^\text{epis} > 0$ are aleatoric and epistemic uncertainties, and $\tilde{\alpha}_i > 0$ is a positive shape score which will later be used to derive a pixel-level shape parameter $\alpha$ in~\Cref{sec:method_pixel}.
We apply sigmoid activation to predict $\bar{c}_i$ as in original NeRF and softplus activation for all other parameters to enforce range restrictions.
The model architecture remains mostly unchanged, except for the three additional output neurons in the last layer to predict $(U_i^\text{alea}, U_i^\text{epis}, \tilde{\alpha}_i)$.

\subsection{Propagation from Points to Pixels}
\label{sec:method_propagation}

Let the set of conditional means and variances of all point colors along a ray be $\boldsymbol{\theta} \coloneq \{ (\mu_i, \sigma_i^2) \}_{i=1}^N$, which fully specifies the conditional distribution of the ray color $c$.
Under this setting, the pixel's predictive mean color, total, aleatoric, and epistemic uncertainties are defined as
\begin{align}
    \bar{c} &\coloneq \E[c],
    &
    U &\coloneq \Var[c],
    \label{eq:quantities_pixel1}
    \\
    U^\text{alea} &\coloneq \E[\Var[c | \boldsymbol{\theta}]],
    &
    U^\text{epis} &\coloneq \Var[\E[c | \boldsymbol{\theta}]].
    \label{eq:quantities_pixel2}
\end{align}

With the point-level predictions from~\Cref{eq:f_evidential} and the independence assumption of points, the pixel color and uncertainties can be derived as
\begin{align}
    \bar{c} &= \sum_{i=1}^N w_i \bar{c}_i,
    &
    U &= \sum_{i=1}^N w_i^2 U_i,
    \label{eq:quantities_pixel_from_point1}
    \\
    U^\text{alea} &= \sum_{i=1}^N w_i^2 U_i^\text{alea},
    &
    U^\text{epis} &= \sum_{i=1}^N w_i^2 U_i^\text{epis}.
    \label{eq:quantities_pixel_from_point2}
\end{align}
In other words, pixel-level aleatoric and epistemic uncertainties can be obtained as weighted sums of their point-level counterparts, where the weights are equal to the squared weights for color volumetric rendering.
Detailed proofs are provided in~\Cref{sec:derivations_prediction_and_uncertainty_propagation}.

\subsection{Pixel-level Probabilistic Radiance Modeling}
\label{sec:method_pixel}

Now we introduce the probabilistic model for pixel colors.
Recall that $\mu = \sum_{i=1}^N w_i \mu_i$ and $\sigma^2 = \sum_{i=1}^N w_i^2 \sigma_i^2$ represent the conditional mean and variance of the pixel color $c$ in~\Cref{eq:pixel_normal}.
To quantify aleatoric and epistemic uncertainties of a pixel color in closed form, we model $(\mu, \sigma^2)$ by a normal-inverse-gamma (NIG) distribution
\begin{equation}
    \label{eq:pixel_nig}
    \mu, \sigma^2 \sim \text{NIG}(\gamma, \nu, \alpha, \beta),
\end{equation}
or equivalently, $\mu \ | \ \sigma^2 \sim \N(\gamma, \sigma^2 / \nu)$ and $\sigma^2 \sim \Gamma^{-1}(\alpha, \beta)$,
where $\gamma \in [0, 1]$, $\nu > 0$, and $\Gamma^{-1}(\alpha, \beta)$ denotes an inverse-gamma distribution with shape $\alpha > 1$ and scale $\beta > 0$.
Consider each pixel color $c$ as being generated through a hierarchical sampling process:
Given the higher-order evidential distribution $\text{NIG}(\gamma, \nu, \alpha, \beta)$, drawing a sample $(\mu, \sigma^2)$ from it yields an instance of lower-order normal distribution $\N(\mu, \sigma^2)$, from which the pixel color $c$ is sampled.

With this formulation, the pixel color’s predictive mean, total, aleatoric, and epistemic uncertainties can be directly expressed in terms of the NIG parameters:
\begin{align}
    \bar{c}
    &= \E[c]
    = \E[\E[c | \mu, \sigma^2]]
    = \E[\mu]
    = \gamma,
    \\
    \label{eq:U}
    U
    &= \Var[c]
    = U^\text{alea} + U^\text{epis},
    \\
    \label{eq:AU}
    U^\text{alea}
    &= \E[\Var[c | \mu, \sigma^2]]
    = \E[\sigma^2]
    = \frac{\beta}{\alpha - 1},
    \\
    \label{eq:EU}
    U^\text{epis}
    &= \Var[\E[c | \mu, \sigma^2]]
    = \Var[\mu]
    = \frac{\beta}{(\alpha - 1)\nu}.
\end{align}

\begin{table*}[h]
    \centering
    \resizebox{\textwidth}{!}{
        \newcolumntype{M}{>{\centering\arraybackslash}m{0.095\textwidth}}
        \begin{tabular}{cc|*{6}{M}}
        \toprule[1pt]
            Dataset
            &Method
            &PSNR$\uparrow$
            &SSIM$\uparrow$
            &LPIPS$\downarrow$
            &NLL$\downarrow$
            &AUSE\newline RMSE$\downarrow$
            &AUSE\newline MAE$\downarrow$
        \\ \midrule
            \multirow{7}{*}{\rotatebox{90}{LF~\cite{lf}}}
            &Baseline
            &28.5538
            &0.9172
            &0.0465
        \\
            &Dropout~\cite{mcdropout}
            &28.1137
            &0.9061
            &0.0569
            &3.6670
            &0.0125
            &0.0049
        \\
            &Normal~\cite{normal}
            &28.0064
            &0.9165
            &0.0531
            &0.4425
            &0.0090
            &\third{0.0029}
        \\
            &MoL~\cite{mixnerf}
            &28.2200
            &0.9095
            &0.0672
            &\first{-2.5393}
            &\third{0.0084}
            &\third{0.0029}
        \\
            &Ensembles~\cite{ensembles}
            &\second{29.3779}
            &\second{0.9308}
            &\second{0.0411}
            &0.3245
            &\first{0.0070}
            &\second{0.0026}
        \\
            &DANE~\cite{dane}
            &\second{29.3779}
            &\second{0.9308}
            &\second{0.0411}
            &\third{-0.4317}
            &0.0101
            &0.0039
        \\
            &Evidential
            &\first{29.9679}
            &\first{0.9345}
            &\first{0.0359}
            &\second{-2.4491}
            &\first{0.0070}
            &\first{0.0025}
        \\ \midrule
            \multirow{7}{*}{\rotatebox{90}{LLFF~\cite{llff}}}
            &Baseline
            &17.5281
            &0.4720
            &0.4119
        \\
            &Dropout~\cite{mcdropout}
            &17.1423
            &0.4441
            &0.4468
            &90.6773
            &0.0742
            &0.0426
        \\
            &Normal~\cite{normal}
            &16.5500
            &0.4135
            &0.4732
            &55.3580
            &0.0674
            &0.0405
        \\
            &MoL~\cite{mixnerf}
            &16.4218
            &0.4237
            &0.4842
            &\second{2.2470}
            &0.0814
            &0.0384
        \\
            &Ensembles~\cite{ensembles}
            &\first{17.9181}
            &\first{0.5109}
            &\second{0.3932}
            &11.1658
            &\first{0.0513}
            &\first{0.0253}
        \\
            &DANE~\cite{dane}
            &\first{17.9181}
            &\first{0.5109}
            &\second{0.3932}
            &\third{9.7273}
            &\second{0.0521}
            &\second{0.0260}
        \\
            &Evidential
            &\third{17.8793}
            &\third{0.5068}
            &\first{0.3751}
            &\first{0.6765}
            &\third{0.0578}
            &\third{0.0295}
        \\ \midrule
            \multirow{7}{*}{\rotatebox{90}{RobustNeRF~\cite{robustnerf}}}
            &Baseline
            &25.2205
            &0.8296
            &0.1577
        \\
            &Dropout~\cite{mcdropout}
            &24.7693
            &0.8118
            &0.1745
            &22.8799
            &0.0284
            &0.0160
        \\
            &Normal~\cite{normal}
            &25.2987
            &0.8522
            &\second{0.1311}
            &10.4873
            &\third{0.0250}
            &\third{0.0151}
        \\
            &MoL~\cite{mixnerf}
            &23.7874
            &0.7921
            &0.2049
            &\first{-1.3947}
            &0.0304
            &0.0176
        \\
            &Ensembles~\cite{ensembles}
            &\second{26.1953}
            &\second{0.8562}
            &\third{0.1438}
            &4.6309
            &\first{0.0164}
            &\first{0.0098}
        \\
            &DANE~\cite{dane}
            &\second{26.1953}
            &\second{0.8562}
            &\third{0.1438}
            &\third{4.1092}
            &0.0283
            &0.0154
        \\
            &Evidential
            &\first{26.2292}
            &\first{0.8641}
            &\first{0.1112}
            &\second{-1.2702}
            &\second{0.0221}
            &\second{0.0138}
        \\ \bottomrule[1pt]
        \end{tabular}
    }
    \caption{
        Quantitative results of scene reconstruction and uncertainty quantification on three datasets, averaged over three independent runs.
        Colored cells denote the \highlightinline{tabfirst}{first}, 
        \highlightinline{tabsecond}{second}, and 
        \highlightinline{tabthird}{third} best results.
        See per-scene statistics with standard deviations in supplementary~\Cref{sec:further_discussions}.
    }
    \vspace{-4mm}
    \label{tab:quantitative}
\end{table*}

Since the Evidential NeRF model directly predicts uncertainties and shape scores instead of NIG parameters, the NIG parameters need to be reformulated as
\vspace{-0.2cm}
\begin{align}
    \gamma &= \bar{c},
    &
    \nu &= \frac{U^\text{alea}}{U^\text{epis}},
    \\
    \alpha &= 1 + \sum_{i=1}^N \tilde{w} \tilde{\alpha}_i,
    &
    \beta &= U^\text{alea}(\alpha - 1),
\end{align}
where $\bar{c}$, $U^\text{alea}$, and $U^\text{epis}$ are obtained from the point-to-pixel propagation process in~\Cref{eq:quantities_pixel_from_point1,eq:quantities_pixel_from_point2}, $\tilde{\alpha}_i$ is the point's shape score, and $\tilde{w}_i \coloneq w_i / \sum_{j=1}^N w_j$ is the normalized weight that determines how much a point's shape score contributes to the NIG shape parameter $\alpha$ of the pixel.

\begin{figure*}[t]
    \centering
    \includegraphics[width=\linewidth]{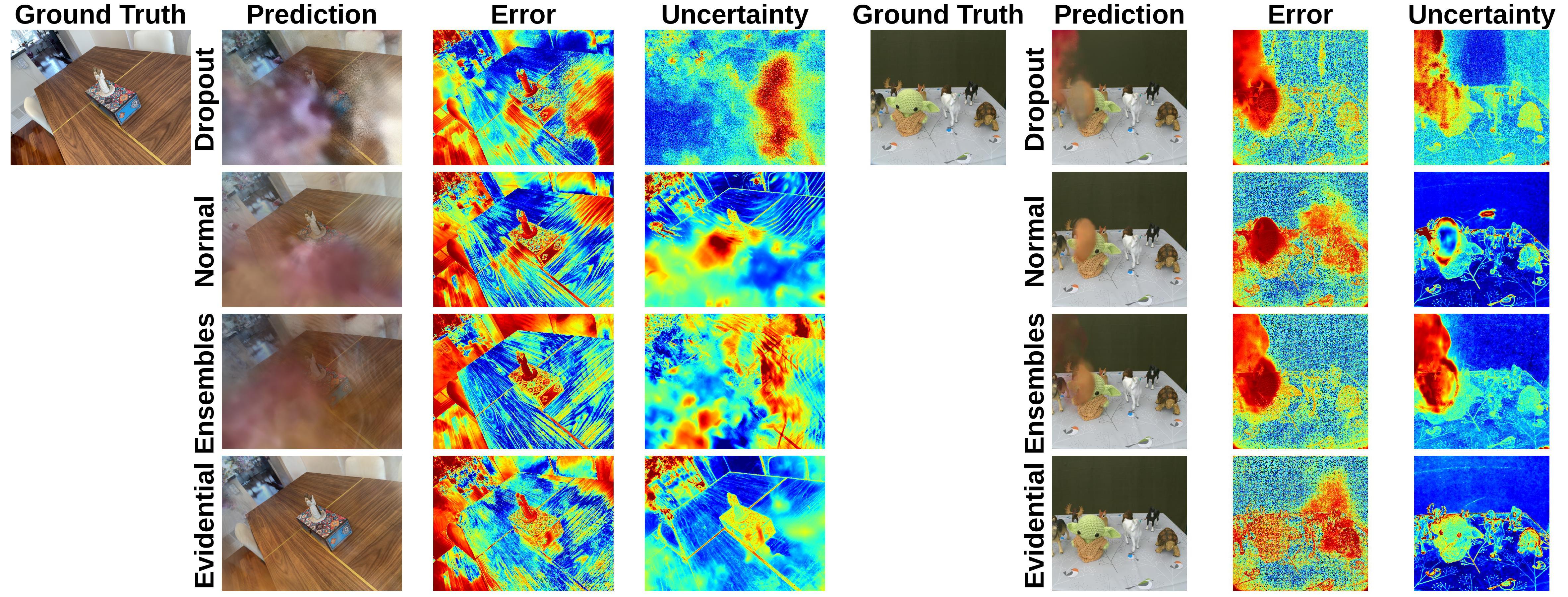}
    \vspace{-6mm}
    \caption{
        Qualitative comparison on two example scenes, with image reconstructions, error maps, and uncertainty maps.
        Histogram equalization is conducted on the error maps to highlight the error regions.
        Our method's uncertainty is total uncertainty.
        In general, our method achieves better reconstruction accuracy and produces uncertainty maps that are more consistent with prediction errors.
    }
    \label{fig:qualitative}
\end{figure*}

\subsection{Learning}
\label{sec:method_learning}

Based on~\Cref{eq:pixel_normal,eq:pixel_nig}, a pixel color $c$ marginally follows a Student's $t$ distribution
\begin{equation}
    \label{eq:marginal}
    c \sim t\left( \gamma, \frac{\beta(\nu+1)}{\alpha\nu}, 2\alpha \right),
\end{equation}
where $t(\mu_t, \sigma_t^2, \nu_t)$ denotes a Student's $t$ distribution with location $\mu_t$, scale $\sigma_t$, and degrees of freedom $\nu_t$.
The model is trained via maximum likelihood estimation by minimizing the negative log-likelihood (NLL) of the ground truth
\begin{align}
    \mathcal{L}_\text{nll}
    &= -\log p(c^\text{gt} | \gamma, \nu, \alpha, \beta)
    \\
    &= \frac{1}{2}\log\frac{\pi}{\nu}
    - \alpha \log\Omega
    + \log\frac{\Gamma(\alpha)}{\Gamma\left(\alpha + \frac{1}{2}\right)}
    \nonumber \\
    &\quad
    + \left(\alpha+\frac{1}{2}\right) \log((c^\text{gt}-\gamma)^2\nu + \Omega)
    \label{eq:L_nll},
\end{align}
where $c^\text{gt} \in [0, 1]$ is the ground truth pixel color, $\Omega = 2\beta(\nu+1)$, and $\Gamma$ represents the Gamma function.
See derivations in~\Cref{sec:pixel_color_distribution_and_loss_function}.

Different values of $\nu$ and $\beta$ may result in the same scale of the $t$ distribution so long as the ratio $\beta(\nu + 1) / \nu$ is fixed. To resolve this ambiguity and also to suppress excessive evidence assigned to inaccurate predictions, a regularizer~\cite{der} is introduced to the loss function as
\begin{equation}
    \label{eq:reg}
    \mathcal{L}_\text{reg} = |c^\text{gt} - \gamma| (2\nu + \alpha),
\end{equation}
where $|c^\text{gt} - \gamma|$ is the absolute error of prediction and $2\nu + \alpha$ is the count of virtual observations representing the evidence~\cite{der}.
Therefore, the total loss is written as
\begin{equation}
    \label{eq:L_edl}
    \mathcal{L} = \mathcal{L}_\text{nll} + \lambda_\text{reg} \mathcal{L}_\text{reg},
\end{equation}
where $\lambda_\text{reg} > 0$ is the regularization coefficient.

Lastly, we generalize our color channel assumption from a single channel to three for RGB color modeling.
We assume that the three color channels have different means but the same uncertainty, due to the high correlation of variance in different color channels of the same pixel~\cite{nerfw,activenerf}.
Practically, the output dimension of color prediction head in~\Cref{eq:f_evidential} is set to three, predicting the point's mean RGB color as $\bar{\boldsymbol{c}}_i \in [0, 1]^3$, and the uncertainty-related parameters are broadcast to all color channels during training.

\begin{figure}[t]
    \centering
    \vspace{-5mm}
    \includegraphics[width=\linewidth]{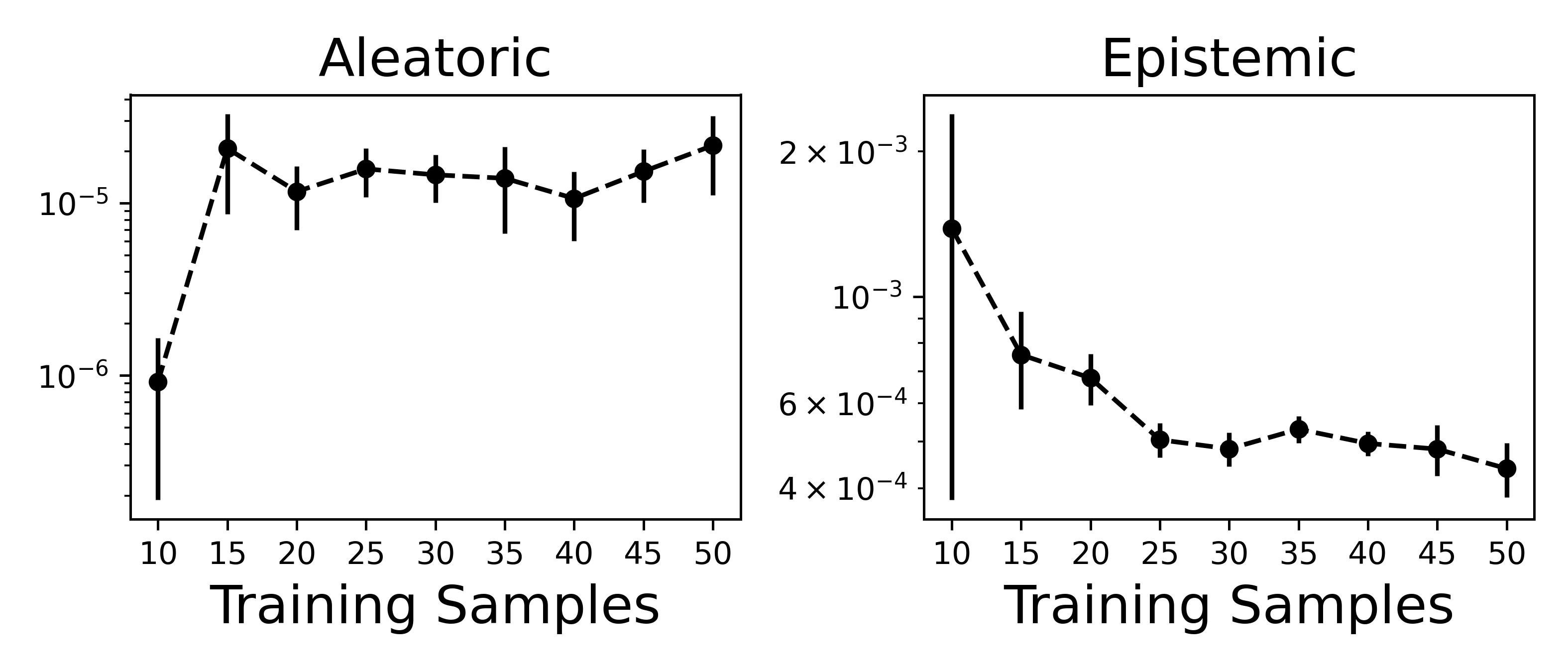}
    \vspace{-8mm}
    \caption{
        Test uncertainties vs. training sample size on \textit{Android}.
    }
    \vspace{-4mm}
\label{fig:uncertainty_scale}
\end{figure}

\section{Experiments}

\subsection{Settings}

\paragraphtitle{Datasets.}
We test the UQ methods on three 3D scene reconstruction datasets: Light Field (LF)~\cite{lf}, Local Light Field Fusion (LLFF)~\cite{llff}, and RobustNeRF~\cite{robustnerf}.
For LF, we test on the four scenes following the setup in~\cite{bayesrays} with the same train-test split.
For LLFF, we employ its eight scenes and adopt the train-test split of~\cite{regnerf}, training with only three input views to assess UQ performance under sparse-view conditions.
For RobustNeRF, we evaluate on its four scenes following their protocol of training on cluttered images only and testing on clean ones.

\paragraphtitle{Metrics.}
We evaluate the method performance based on the accuracy of both the rendered images and the uncertainty estimates.
For images, we report PSNR, SSIM, and LPIPS to reflect the image reconstruction quality.
For uncertainties, we use negative log-likelihood (NLL) and area under sparsification error (AUSE) with respect to both RMSE and MAE, measuring the quality of uncertainty estimates in terms of distributional fit and error ranking capabilities.

\paragraphtitle{Baselines.}
We select UQ methods from each of the three categories discussed in~\Cref{sec:related_work} as baselines.
For closed-form likelihood models, we include the normal distribution approach~\cite{normal} adopted in~\cite{nerfw} and~\cite{activenerf} and mixture of Laplace distributions (MoL) employed in~\cite{mixnerf} and~\cite{flipnerf}.
Among Bayesian methods, we choose the classical Monte Carlo dropout approach~\cite{mcdropout}.
For ensemble methods, we consider naive deep ensembles~\cite{ensembles} and density-aware NeRF ensembles (DANE)~\cite{dane}.

\begin{table}[t]
    \centering
    \resizebox{\columnwidth}{!}{
        \newcolumntype{M}{>{\centering\arraybackslash}c{0.095\columnwidth}}
        \begin{tabular}{c | c | *{6}{c}}
        \toprule[1pt]
            Mode
            &Baseline
            &Dropout
            &Normal
            &MoL
            &Ens.
            &DANE
            &Ours
        \\ \midrule
            Train/min.$\downarrow$
            & 11.84
            & 88.54
            & \first{12.22}
            & \second{12.71}
            & 59.22
            & 59.22
            & \third{13.57}
        \\
            Infer/FPS$\uparrow$
            & 4.88
            & 0.09
            & \first{4.71}
            & \third{4.42}
            & 0.96
            & 0.96
            & \second{4.67}
        \\ \bottomrule[1pt]
        \end{tabular}
    }
    \vspace{-2mm}
    \caption{
        Average training time per $30k$ steps and inference FPS of baseline nerfacto and different UQ methods on an A6000 GPU.
    }
    \vspace{-0.4cm}
    \label{tab:time}
\end{table}

\paragraphtitle{Implementation.}
Prior benchmarks on NeRF uncertainty quantification suffer from varying choices of data splits, model architectures, and training schemes.
To isolate UQ method effects from engineering confounders, we establish a new standardized benchmark to focus comparison on the underlying UQ approaches themselves.
First, for data split, we follow the aforementioned scheme for all models.
Second, for architecture, we use nerfacto model in~\cite{nerfstudio} to implement all methods for efficiency.
For likelihood models, the only difference is the output layer size and the loss function.
For the Bayesian method, Monte Carlo dropout is implemented with a dropout probability $0.2$, trained once and sampled five times at inference.
For ensembles, both naive ensembles and DANE are implemented by training five models independently.
Third, in training, we adopt the default batch size, optimizer, and learning rate scheduler of nerfacto for all methods and train all models for the same number of iterations, chosen based on the convergence speed on each scene.
For stability, we run each method independently three times and report the averaged metrics.

\subsection{Results}

\paragraphtitle{Quantitative results.}
The quantitative performance of the uncertainty quantification methods is detailed in \Cref{tab:quantitative} with per-scene statistics provided in \Cref{sec:further_discussions}.

\begin{figure}[t]
    \centering
    \includegraphics[width=\linewidth]{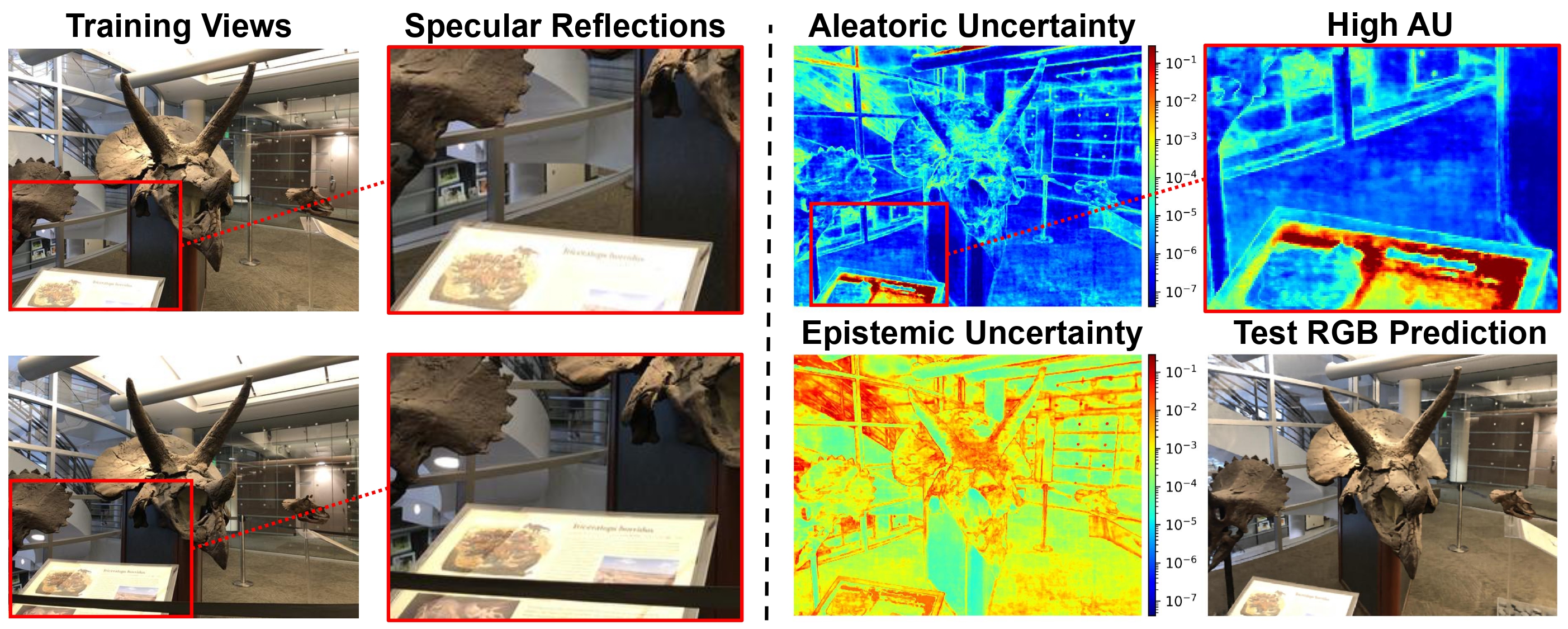}
    \vspace{-6mm}
    \caption{
        A case where AU dominates EU.
        The highly reflective surface of the display case in the foreground incurs specular reflections.
        AU arises due to the presence of data noise caused by the inconsistency of light across different training views.
    }
    \vspace{-2mm}
    \label{fig:specular}
\end{figure}

For \textbf{scene reconstruction}, compared to the baseline nerfacto model, likelihood and Bayesian methods (Normal, MoL, Dropout) struggle to preserve the image reconstruction accuracy, indicating that their predictive performance is compromised in exchange for uncertainty estimation.
Ensemble-based methods provide faithful reconstructions in general, but they are extremely inefficient due to their substantial computational cost in both training and inference.
In contrast, Evidential NeRF, while being efficient, consistently outperforms the baseline in image reconstruction, demonstrating that our uncertainties need not come at the expense of rendering fidelity.
Remarkably, with only a single network trained and a single forward pass required at inference, the evidential approach achieves prediction accuracy comparable to, and in $7$ out of $9$ image reconstruction metrics even surpassing, the computationally expensive ensemble methods.

For \textbf{uncertainty quantification}, in terms of NLL, MoL model achieves the best result overall likely due to its multimodal radiance distribution assumption.
Compared to the normal distribution with fixed mean and variance, a normal distribution with random mean and variance increases test data likelihood by factors of $1.8 \times 10^1$, $5.6 \times 10^{23}$, and $1.3 \times 10^5$ on the three datasets, respectively.
This tremendous improvement is because the fixed mean and variance assumption used by classical normal-based UQ only captures the aleatoric variations within the training data while overlooking epistemic uncertainty.
The likelihood gain is less pronounced on LF because its test images are adjacent to the training views, thus the distribution shift is relatively insignificant.
As for AUSE, ensemble-based methods achieve the strongest results in general, thanks to their ability to represent predictive variability through multiple models.
Nevertheless, our evidential approach remains highly competitive, often ranking second only to ensembles, demonstrating that jointly modeling aleatoric and epistemic factors yields uncertainty estimates with stronger correlation with errors.

\paragraphtitle{Qualitative results.}
We qualitatively compare our Evidential NeRF against other methods in~\Cref{fig:qualitative} and \Cref{sec:further_discussions}.
A primary goal of UQ is to produce uncertainty maps that accurately identify potential model failures. 
Previous methods, by neglecting either the aleatoric or epistemic component of predictive uncertainty, often fall short of accurately delimiting regions of predictive inaccuracy.
In contrast, by considering both components, our method demonstrates superior alignment with the reconstruction error maps.
This suggests that jointly modeling data noise and the model knowledge gap is essential for generating the most reliable uncertainty estimates.

\paragraphtitle{Uncertainty scaling with data.}
We investigate the effect of training sample size on the magnitude of aleatoric and epistemic uncertainties using an example scene from RobustNeRF in~\Cref{fig:uncertainty_scale}.
We reserve a held-out set of views for evaluation and incrementally select $10$ to $50$ images from the remaining data for training.
As the training sample size increases, the test AU increases while the test EU decreases in general, indicating that the additional, potentially more cluttered observations introduce greater data variability, whereas the model's lack of knowledge is gradually mitigated with more training data.

\begin{figure}[t]
    \centering
    \includegraphics[width=\linewidth]{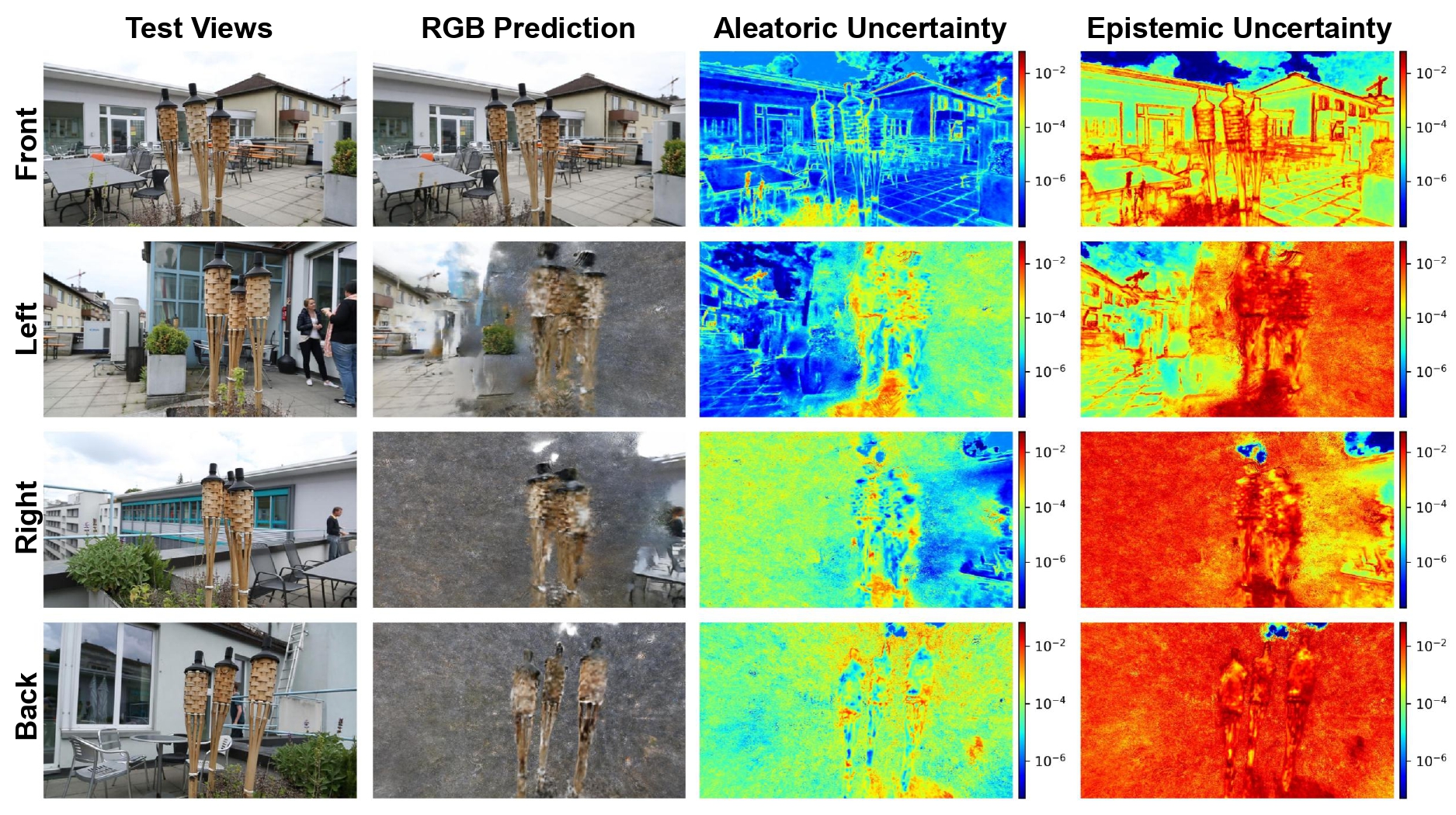}
    \vspace{-6mm}
    \caption{
        A case where EU dominates AU.
        The model is trained only by $5$ images from the front and asked to render the scene from all the viewing angles.
        EU arises due to the lack of knowledge during training on the views out of the training distribution.
    }
    \label{fig:ood}
    \vspace{-4mm}
\end{figure}

\paragraphtitle{Aleatoric and epistemic factors in 3D scenes.}
Several scenarios where aleatoric and epistemic uncertainties emerge are respectively presented in \Cref{fig:teaser,fig:specular,fig:ood}.
\Cref{fig:teaser} illustrates an in-the-wild scene with several aleatoric factors (varying illumination, high-frequency edge regions, transient objects) and an epistemic factor (partial occlusions).
\Cref{fig:specular} showcases a scenario with a reflective object, where aleatoric light variation is the dominant source of potential rendering error.
\Cref{fig:ood} demonstrates an out-of-distribution example where the model is asked to render the scene from angles that never appeared during training, showing that the lack of model knowledge about these unseen views is the primary cause of predictive failure.

\paragraphtitle{Computational efficiency.}
We report the total training time and inference frames per second (FPS) of baseline nerfacto and different UQ methods in~\Cref{tab:time}.
Compared to other UQ methods, in training, our evidential approach is only slightly slower than other two likelihood models while being significantly faster than the ensemble methods.
As for inference, since our approach directly predicts uncertainty fields, it is more efficient than the methods that require computing uncertainty from the predicted parameters, such as MoL.
Since nerfacto runs on Tiny CUDA Neural Networks~\cite{tiny-cuda-nn}, which does not support dropout, dropout is implemented without acceleration, making it naturally slower than other methods.

\begin{figure}[t]
    \centering
    \includegraphics[width=\linewidth]{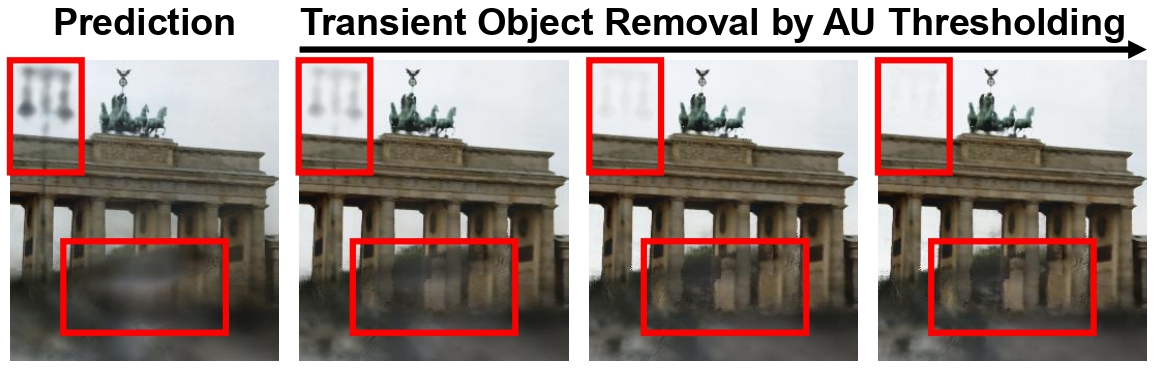}
    \vspace{-6mm}
    \caption{
        Scene cleaning based on aleatoric uncertainty as a post-processing step for floater removal.
        Points with AU above a certain threshold have their density reduced to become more transparent.
        By reducing the threshold, more artifacts can be eliminated.
    }
    \vspace{-0.4cm}
    \label{fig:clean}
\end{figure}

\subsection{Applications}
\begin{figure}[t]
    \centering
    \includegraphics[width=\linewidth]{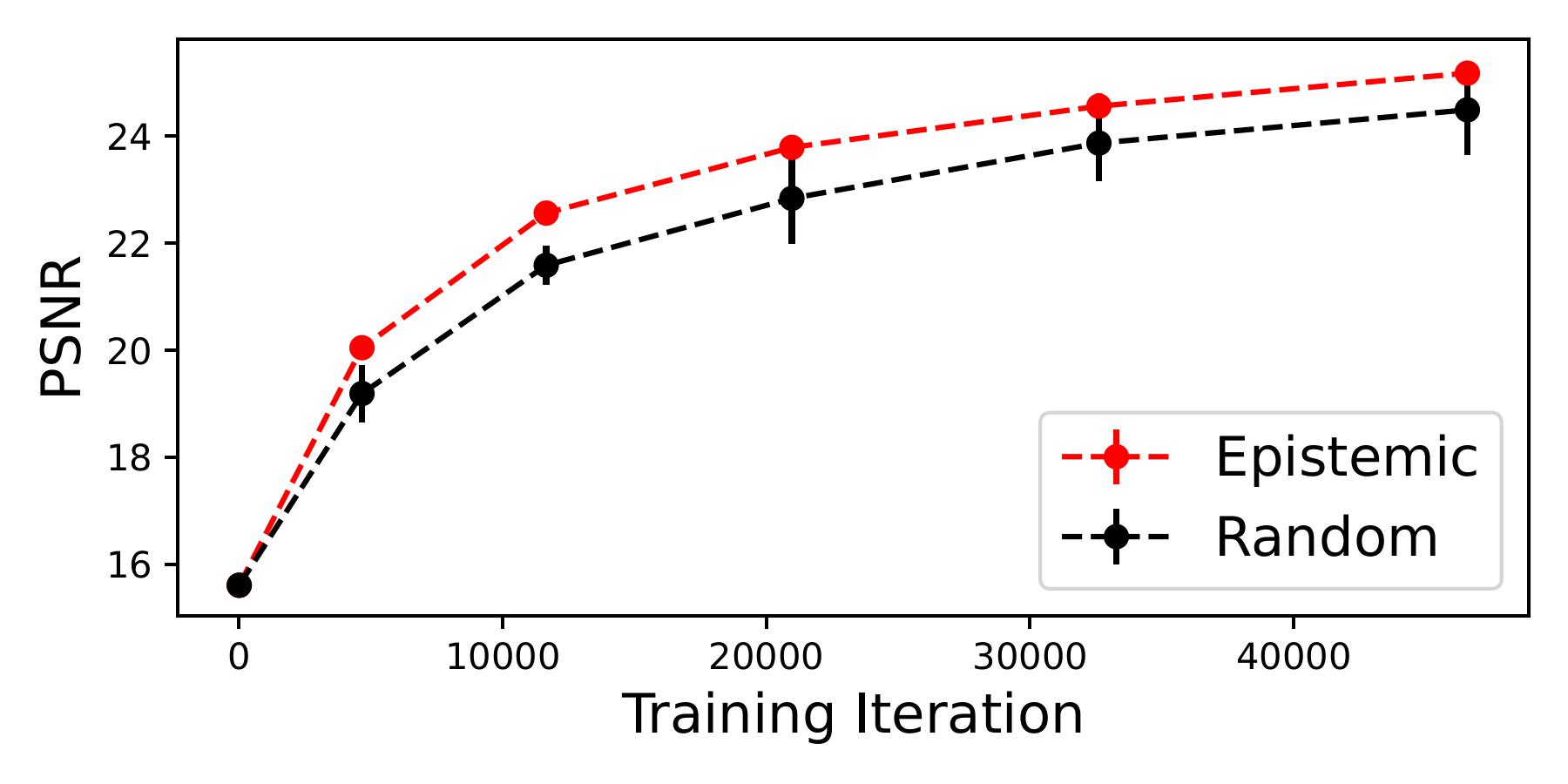}
    \vspace{-8mm}
    \caption{
        Mean and standard derivation of test PSNR of three runs on \textit{Horns} scene, with two active sampling strategies: EU-based selection and random selection.
        The samples identified via epistemic uncertainty are more informative for model learning.
    }
    \label{fig:active}
    \vspace{-0.4cm}
\end{figure}

We demonstrate two example applications using aleatoric and epistemic uncertainties respectively in~\Cref{fig:clean,fig:active}.

\paragraphtitle{Scene cleaning.}
Training NeRFs on unconstrained images often leads to inaccurate predictions due to radiance inconsistencies from uncontrolled lighting changes or transient objects, which are irreducible data noise explained by aleatoric uncertainty.
As a robust indicator of geometric artifacts, AU-based scene cleaning can be applied as a post-processing procedure to eliminate noise in the renderings.
\Cref{fig:clean} presents an example where the erroneous floaters of the rendering can be progressively removed by gradually changing the AU threshold to suppress point densities.

\paragraphtitle{Active learning.}
As a proxy of lack of model knowledge, epistemic uncertainty is an ideal metric for active learning.
\Cref{fig:active} illustrates a next-best-view planning experiment on an LLFF scene.
Starting from an initial training set of $5$ images, an Evidential NeRF is trained iteratively by $5$ rounds.
In each round, the model is trained by $5$ epochs with all the current training images and then tested on a holdout test set, before choosing $5$ additional images from the remaining data to add to the training pool.
Two sampling strategies for data selection are compared: choosing the images with the most EU and choosing images randomly.
The average and standard deviation of three independent runs of each selection scheme are reported.
Active selection based on EU gives noticeably higher PSNR than random selection, indicating the samples with higher epistemic uncertainty are more informative for model learning.

\section{Conclusion}

Despite the efficacy of NeRFs in photorealistic 3D scene reconstruction, the absence of robust predictive uncertainty quantification significantly hinders their deployment in safety-critical domains.
While some uncertainty quantification methods have been proposed for NeRFs, none of them provides a mechanism for quantifying both aleatoric and epistemic uncertainties of the scene reconstructions.
In this paper, we introduce Evidential Neural Radiance Fields, a principled probabilistic framework that resolves this critical gap.
Our approach seamlessly adapts evidential deep learning to the hierarchical nature of NeRFs through a point-to-pixel aleatoric and epistemic uncertainty propagation paradigm.
Through extensive experiments on three standardized benchmarks, Evidential NeRF demonstrates its superiority in both image reconstruction fidelity and uncertainty estimation quality compared to other methods, while identifying the aleatoric and epistemic factors that contribute to predictive uncertainty within a variety of 3D scenes.

\paragraphtitle{Limitation.}
For tractable uncertainty propagation, Evidential NeRF assumes deterministic volumetric density, leaving spatial uncertainty not explicitly modeled.
Extending the formulation to capture the uncertainty in scene geometries would enable more comprehensive uncertainty modeling.

\section{Acknowledgment}
This work is supported by NSF 2112562 Athena AI Institute.

{
    \small
    \bibliographystyle{ieeenat_fullname}
    \bibliography{main}
}

\clearpage
\maketitlesupplementary

\section{Derivations: Point-to-Pixel Propagation of Radiance and Uncertainty}
\label{sec:derivations_prediction_and_uncertainty_propagation}

This section derives the propagation formulae of radiance and uncertainties specified in~\Cref{eq:quantities_pixel_from_point1,eq:quantities_pixel_from_point2}.

The mean color aggregation formula follows directly from the linearity of expectation, which does not require any independence assumption.
\begin{align}
    \bar{c}
    = \E[c]
    = \E \left[\sum_{i=1}^N w_i c_i \right]
    = \sum_{i=1}^N w_i \E[c_i]
    = \sum_{i=1}^N w_i \bar{c}_i.
\end{align}

Deriving the three uncertainty propagation formulae each requires its own independence assumption.
However, only two, and any two, of these three assumptions are necessary to derive all the three formulae.

\begin{assumption}
\label{assumption:1}
The point colors are independent, i.e., $c_i \indep c_j, \ \forall i \neq j$.
\end{assumption}

\begin{assumption}
\label{assumption:2}
The point colors are conditionally independent given the conditional means and variances, i.e., $c_i \indep c_j \mid \boldsymbol{\theta}, \ \forall i \neq j$.
\end{assumption}

\begin{assumption}
\label{assumption:3}
The point colors’ conditional means are independent, i.e., $\mu_i \indep \mu_j, \ \forall i \neq j$.
\end{assumption}

% \begin{assumption}
% \label{assumption:4}
% The conditional distribution of each point color depends only on its own conditional mean and variance, i.e., $c_i \mid \boldsymbol{\theta} \;\overset{d}{=}\; c_i \mid \mu_i, \sigma_i^2, \ \forall i$, where $\overset{d}{=}$ denotes equality in distribution.
% \end{assumption}

With these assumptions, it can be shown that
\begin{align}
    \label{eq:appendix_U}
    U
    &= \Var[c]
    = \Var\left[ \sum_{i=1}^N w_i c_i \right]
    \overset{\text{A.1}}{=} \sum_{i=1}^N w_i^2 \Var[c_i]
    = \sum_{i=1}^N w_i^2 U_i,
    \\
    \label{eq:appendix_AU}
    U^\text{alea}
    &= \E[\Var[c \,|\, \boldsymbol{\theta}]]
    = \E\left[\Var\left[ \sum_{i=1}^N w_i c_i \ \middle|\ \boldsymbol{\theta} \right]\right]
    \overset{\text{A.2}}{=} \E\left[ \sum_{i=1}^N w_i^2 \Var[c_i \,|\, \boldsymbol{\theta}] \right]
    \nonumber
    \\
    &= \sum_{i=1}^N w_i^2 \E\left[ \Var[c_i \,|\, \boldsymbol{\theta}] \right]
    % \overset{\text{A.4}}{=}
    = \sum_{i=1}^N w_i^2 \E\left[ \Var[c_i \,|\, \mu_i, \sigma_i^2] \right]
    = \sum_{i=1}^N w_i^2 U_i^\text{alea},
    \\
    \label{eq:appendix_EU}
    U^\text{epis}
    &= \Var[\E[c \,|\, \boldsymbol{\theta}]]
    = \Var\left[\E\left[\sum_{i=1}^N w_i c_i \ \middle|\ \boldsymbol{\theta} \right]\right]
    = \Var\left[ \sum_{i=1}^N w_i \E[c_i \,|\, \boldsymbol{\theta}] \right]
    \nonumber
    \\
    &\overset{\text{A.3}}{=} \sum_{i=1}^N w_i^2 \Var\left[ \E[c_i \,|\, \boldsymbol{\theta}] \right]
    % \overset{\text{A.4}}{=}
    = \sum_{i=1}^N w_i^2 \Var[\E[c_i \,|\, \mu_i, \sigma_i^2]]
    = \sum_{i=1}^N w_i^2 U_i^\text{epis},
\end{align}
where $\overset{\text{A.}n}{=}$ denotes the step where Assumption $n$ is used.

These three equations are connected by the law of total variance, as
\begin{align}
    \underbrace{\Var[c]}_U
    &= \underbrace{\E[\Var[c \,|\, \boldsymbol{\theta}]]}_{U^\text{alea}}
    + \underbrace{\Var[\E[c \,|\, \boldsymbol{\theta}]]}_{U^\text{epis}},
    \\
    \underbrace{\Var[c_i]}_{U_i}
    &= \underbrace{\E[\Var[c_i \,|\, \mu_i, \sigma_i^2]]}_{U_i^\text{alea}}
    + \underbrace{\Var[\E[c_i \,|\, \mu_i, \sigma_i^2]]}_{U_i^\text{epis}}.
\end{align}
Therefore, any two of~\Cref{eq:appendix_U,eq:appendix_AU,eq:appendix_EU} imply the third, and thus only two of~\Cref{assumption:1,assumption:2,assumption:3} are necessary to derive all the three equations.

\section{Derivations: Pixel Radiance Marginal Distribution and Loss Function}
\label{sec:pixel_color_distribution_and_loss_function}

This section derives the marginal distribution of pixel radiance and the negative log-likelihood loss in~\Cref{eq:marginal,eq:L_nll}.

Evidential NeRF defines each pixel radiance $c$ by a hierarchical probabilistic model
\begin{equation}
    \label{eq:evidential_nerf}
    c \ | \ \mu, \sigma^2 \sim \mathcal{N}(\mu, \sigma^2),
    \qquad\qquad
    \mu \ | \ \sigma^2 \sim \mathcal{N} (\gamma, \sigma^2 / \nu ),
    \qquad\qquad
    \sigma^2 \sim \Gamma^{-1}(\alpha, \beta),
\end{equation}
where $\mathcal{N}(\cdot, \cdot)$ and $\Gamma^{-1}(\cdot, \cdot)$ respectively denote normal distribution and inverse-gamma distribution and $\gamma \in \R$, $\nu > 0$, $\alpha > 1$, and $\beta > 0$ are the evidential NIG parameters.
Based on these prerequisites, we derive the marginal distribution and negative log-likelihood of the pixel color $c$.

The probability density functions of the distributions in~\Cref{eq:evidential_nerf} are respectively given by
\begin{align}
p(c \ | \ \mu, \sigma^2) &= \frac{1}{\sqrt{2\pi\sigma^2}} \exp\left( -\frac{(c - \mu)^2}{2\sigma^2} \right),
\\
p(\mu \ | \ \sigma^2) &= \frac{\sqrt{\nu}}{\sqrt{2\pi\sigma^2}} \exp\left( -\frac{(\mu - \gamma)^2 \nu}{2\sigma^2} \right),
\\
p(\sigma^2) &= \frac{\beta^\alpha}{\Gamma(\alpha)} (\sigma^2)^{-\alpha-1}\exp\left(-\frac{\beta}{\sigma^2}\right).
\end{align}
It can therefore be shown that
\begingroup
\allowdisplaybreaks
\begin{align}
    &\quad\,\,\,\, p(c \ | \ \sigma^2)
    \\
    &= \int_{-\infty}^{\infty} p(c \ | \ \mu, \sigma^2) p(\mu \ | \ \sigma^2) d\mu
    \\
    &= \int_{-\infty}^{\infty}
    \frac{1}{\sqrt{2\pi\sigma^2}} \exp\left( -\frac{(c - \mu)^2}{2\sigma^2} \right)
    \frac{\sqrt{\nu}}{\sqrt{2\pi\sigma^2}} \exp\left( -\frac{(\mu - \gamma)^2 \nu}{2\sigma^2} \right)
    d\mu
    \\
    &= \int_{-\infty}^{\infty}
    \frac{\sqrt{\nu}}{2\pi\sigma^2} \exp\left( -\frac{(c - \mu)^2 + (\mu - \gamma)^2 \nu}{2\sigma^2} \right)
    d\mu
    \\
    &= \int_{-\infty}^{\infty}
    \frac{\sqrt{\nu}}{2\pi\sigma^2} \exp\left( -\frac{
    (\nu+1)\mu^2 -2(c+\gamma\nu)\mu + (c^2+\gamma^2\nu)
    }{2\sigma^2} \right)
    d\mu
    \\
    &= \frac{\sqrt{\nu}}{2\pi\sigma^2}
    \exp\left(
    -\frac{c^2+\gamma^2\nu}{2\sigma^2}
    \right)
    \int_{-\infty}^{\infty}
    \exp\left( -\frac{
    (\nu+1)\mu^2 -2(c+\gamma\nu)\mu
    }{2\sigma^2} \right)
    d\mu
    \\
    &= \frac{\sqrt{\nu}}{2\pi\sigma^2}
    \exp\left(
    -\frac{c^2+\gamma^2\nu}{2\sigma^2}
    \right)
    \int_{-\infty}^{\infty}
    \exp\left(
    -\frac{(\mu-\frac{c+\gamma\nu}{\nu+1})^2 -(\frac{c+\gamma\nu}{\nu+1})^2}
    {\frac{2\sigma^2}{\nu+1}}
    \right)
    d\mu
    \\
    &= \frac{\sqrt{\nu}}{2\pi\sigma^2}
    \exp\left(
    -\frac{c^2+\gamma^2\nu}{2\sigma^2}
    \right)
    \exp\left(
    \frac{(c+\gamma\nu)^2}{2\sigma^2(\nu+1)}
    \right)
    \int_{-\infty}^{\infty}
    \exp\left(
    -\frac{(\mu-\frac{c+\gamma\nu}{\nu+1})^2}{\frac{2\sigma^2}{\nu+1}}
    \right)
    d\mu
    \\
    &= \frac{\sqrt{\nu}}{2\pi\sigma^2}
    \exp\left(
    -\frac{c^2+\gamma^2\nu}{2\sigma^2}
    +\frac{(c+\gamma\nu)^2}{2\sigma^2(\nu+1)}
    \right)
    \int_{-\infty}^{\infty}
    \sqrt{2\pi \frac{\sigma^2}{\nu+1}}
    \mathcal{N} \left(
    \mu ;
    \frac{c+\gamma\nu}{\nu+1},
    \frac{\sigma^2}{\nu+1}
    \right)
    d\mu
    \\
    &= \frac{\sqrt{\nu}}{2\pi\sigma^2}
    \sqrt{2\pi \frac{\sigma^2}{\nu+1}}
    \exp\left(
    -\frac{(c^2+\gamma^2\nu)(\nu+1) - (c+\gamma\nu)^2}{2\sigma^2(\nu+1)}
    \right)
    \int_{-\infty}^{\infty}
    \mathcal{N} \left(
    \mu ;
    \frac{c+\gamma\nu}{\nu+1},
    \frac{\sigma^2}{\nu+1}
    \right)
    d\mu
    \\
    &= \frac{\sqrt{\nu}}{2\pi\sigma^2}
    \sqrt{2\pi \frac{\sigma^2}{\nu+1}}
    \exp\left(
    -\frac{(c^2+\gamma^2\nu)(\nu+1) - (c+\gamma\nu)^2}{2\sigma^2(\nu+1)}
    \right)
    \\
    &= \sqrt{\frac{\nu}{2\pi\sigma^2(\nu+1)}}
    \exp\left(
    -\frac{
    (c-\gamma)^2\nu
    }{2\sigma^2(\nu+1)}
    \right),
\end{align}
\endgroup
i.e.,
\begin{equation}
    c \ | \ \sigma^2 \sim \mathcal{N}\left(\gamma, \frac{\sigma^2(\nu+1)}{\nu}\right),
\end{equation}
\newpage
and thus,
\begingroup
\allowdisplaybreaks
\begin{align}
    &\quad\,\,\,\, p(c)
    \\
    &= \int_{0}^{\infty} p(c \ | \ \sigma^2) p(\sigma^2) d\sigma^2
    \\
    &= \int_{0}^{\infty}
    \sqrt{\frac{\nu}{2\pi\sigma^2(\nu+1)}}
    \exp\left(
    -\frac{
    (c-\gamma)^2\nu
    }{2\sigma^2(\nu+1)}
    \right)
    \frac{\beta^\alpha}{\Gamma(\alpha)} (\sigma^2)^{-\alpha-1}\exp\left(-\frac{\beta}{\sigma^2}\right)
    d\sigma^2
    \\
    &=
    \sqrt{\frac{\nu}{2\pi(\nu+1)}}
    \frac{\beta^\alpha}{\Gamma(\alpha)}
    \int_{0}^{\infty}
    \exp\left(
    -\frac{
    (c-\gamma)^2\nu
    }{2\sigma^2(\nu+1)}
    -\frac{\beta}{\sigma^2}
    \right)
    (\sigma^2)^{-\alpha-\frac{3}{2}}
    d\sigma^2
    \\
    &=
    \sqrt{\frac{\nu}{2\pi(\nu+1)}}
    \frac{\beta^\alpha}{\Gamma(\alpha)}
    \int_{0}^{\infty}
    \exp\left(
    -\frac{\frac{(c-\gamma)^2\nu}{2(\nu+1)}+\beta}{\sigma^2}
    \right)
    (\sigma^2)^{-\alpha-\frac{3}{2}}
    d\sigma^2
    \\
    &=
    \sqrt{\frac{\nu}{2\pi(\nu+1)}}
    \frac{\beta^\alpha}{\Gamma(\alpha)}
    \int_{0}^{\infty}
    \frac{\Gamma\left(\alpha+\frac{1}{2}\right)}{\left(\frac{(c-\gamma)^2\nu}{2(\nu+1)}+\beta\right)^{\alpha+\frac{1}{2}}}
    \Gamma^{-1}\left(
    \sigma^2;
    \alpha+\frac{1}{2},
    \frac{(c-\gamma)^2\nu}{2(\nu+1)}+\beta
    \right)
    d\sigma^2
    \\
    &=
    \sqrt{\frac{\nu}{2\pi(\nu+1)}}
    \frac{\beta^\alpha}{\Gamma(\alpha)}
    \frac{\Gamma\left(\alpha+\frac{1}{2}\right)}{\left(\frac{(c-\gamma)^2\nu}{2(\nu+1)}+\beta\right)^{\alpha+\frac{1}{2}}}
    \int_{0}^{\infty}
    \Gamma^{-1}\left(
    \sigma^2;
    \alpha+\frac{1}{2},
    \frac{(c-\gamma)^2\nu}{2(\nu+1)}+\beta
    \right)
    d\sigma^2
    \\
    &=
    \sqrt{\frac{\nu}{2\pi(\nu+1)}}
    \frac{\beta^\alpha}{\Gamma(\alpha)}
    \frac{\Gamma\left(\alpha+\frac{1}{2}\right)}{\left(\frac{(c-\gamma)^2\nu}{2(\nu+1)}+\beta\right)^{\alpha+\frac{1}{2}}}
    \\
    &=
    \frac{\Gamma\left(\alpha+\frac{1}{2}\right)}{\Gamma(\alpha)\sqrt{2\pi\frac{\beta(\nu+1)}{\nu}}}
    \left(\frac{(c-\gamma)^2}{2\frac{\beta(\nu+1)}{\nu}}+1\right)^{-\left(\alpha+\frac{1}{2}\right)}
    \\
    &=
    \frac{\Gamma(\frac{\nu_t+1}{2})}{\Gamma(\frac{\nu_t}{2})\sqrt{\pi\nu_t \sigma_t^2}}
    \left(\frac{(c-\mu_t)^2}{\nu_t \sigma_t^2}+1\right)^{-\frac{\nu_t+1}{2}},
\end{align}
\endgroup
i.e.,
\begin{equation}
    c \sim t\left(
    \mu_t = \gamma,
    \sigma_t^2 = \frac{\beta(\nu+1)}{\alpha\nu},
    \nu_t = 2\alpha
    \right),
\end{equation}
where $t(\mu_t, \sigma_t^2, \nu_t)$ denotes a Student's $t$ distribution with location $\mu_t$, scale $\sigma_t$, and degrees of freedom $\nu_t$, and $\mathcal{N}(x;\cdot,\cdot)$ and $\Gamma^{-1}(x;\cdot,\cdot)$ represent their probability distribution densities at $x$.
This yields \Cref{eq:marginal}.

The negative log-likelihood of $c$ can thereby be derived as
\begingroup
\allowdisplaybreaks
\begin{align}
    &\quad -\log p(c)
    \\
    &= -\log 
    \left(
    \frac{\Gamma\left(\alpha+\frac{1}{2}\right)}{\Gamma(\alpha)\sqrt{2\pi\frac{\beta(\nu+1)}{\nu}}}
    \left(\frac{(c-\gamma)^2}{2\frac{\beta(\nu+1)}{\nu}}+1\right)^{-\left(\alpha+\frac{1}{2}\right)}
    \right)
    \\
    &=
    \frac{1}{2} \log \left( 2\pi\frac{\beta(\nu+1)}{\nu} \right)
    +\log \frac{\Gamma(\alpha)}{\Gamma\left(\alpha+\frac{1}{2}\right)}
    +\left(\alpha+\frac{1}{2}\right) \log\left(\frac{(c-\gamma)^2}{2\frac{\beta(\nu+1)}{\nu}}+1\right)
    \\
    &=
    \frac{1}{2} \log \frac{\pi}{\nu}
    + \frac{1}{2} \log \left( 2\beta(\nu+1) \right)
    +\log \frac{\Gamma(\alpha)}{\Gamma\left(\alpha+\frac{1}{2}\right)}
    +\left(\alpha+\frac{1}{2}\right) \log\left(\frac{(c-\gamma)^2\nu}{2\beta(\nu+1)}+1\right)
    \\
    &=
    \frac{1}{2} \log \frac{\pi}{\nu}
    -\alpha \log \left( 2\beta(\nu+1) \right)
    +\log \frac{\Gamma(\alpha)}{\Gamma\left(\alpha+\frac{1}{2}\right)}
    +\left(\alpha+\frac{1}{2}\right) \log\left((c-\gamma)^2\nu + 2\beta(\nu+1) \right)
    \\
    &=
    \frac{1}{2} \log \frac{\pi}{\nu}
    -\alpha \log \Omega 
    +\log \frac{\Gamma(\alpha)}{\Gamma\left(\alpha+\frac{1}{2}\right)}
    +\left(\alpha+\frac{1}{2}\right) \log\left((c-\gamma)^2\nu + \Omega \right),
\end{align}
\endgroup
where $\Omega = 2\beta(1+\nu)$.
This yields \Cref{eq:L_nll}.

\section{Further Discussions}
\label{sec:further_discussions}

\paragraphtitle{More quantitative results.}
In the main paper, quantitative results averaged across all scenes within each dataset are reported.
\Cref{tab:quantitative_lf,tab:quantitative_llff1,tab:quantitative_llff2,tab:quantitative_robustnerf} provide per-scene statistics including the mean and standard deviation of metrics over three independent runs.

\begin{figure*}[t]
    \centering
    \includegraphics[width=\textwidth]{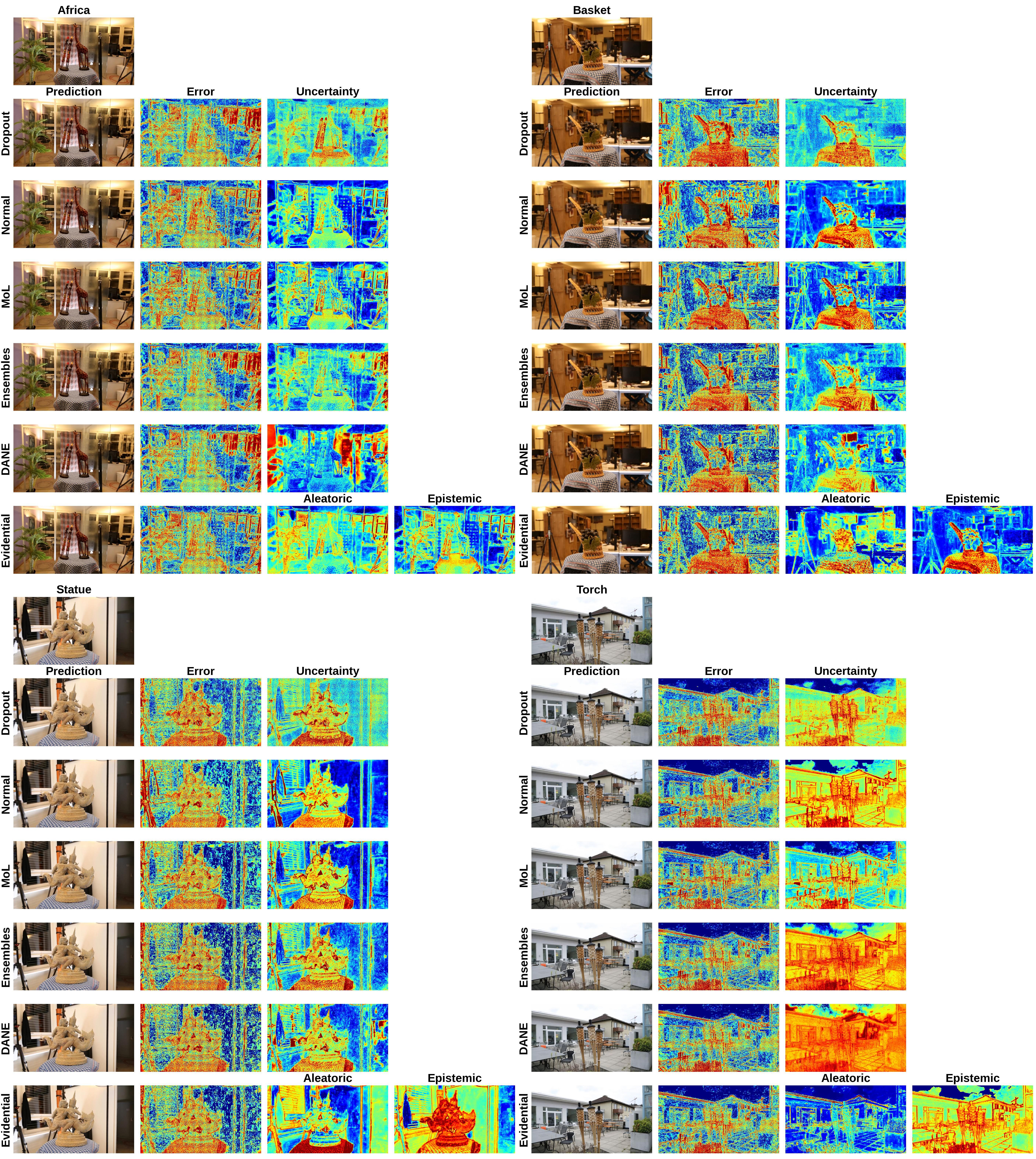}
    \caption{
        Qualitative comparison on LF.
    }
    \label{fig:qualitative_lf}
\end{figure*}

\begin{figure*}[t]
    \centering
    \includegraphics[height=\textheight]{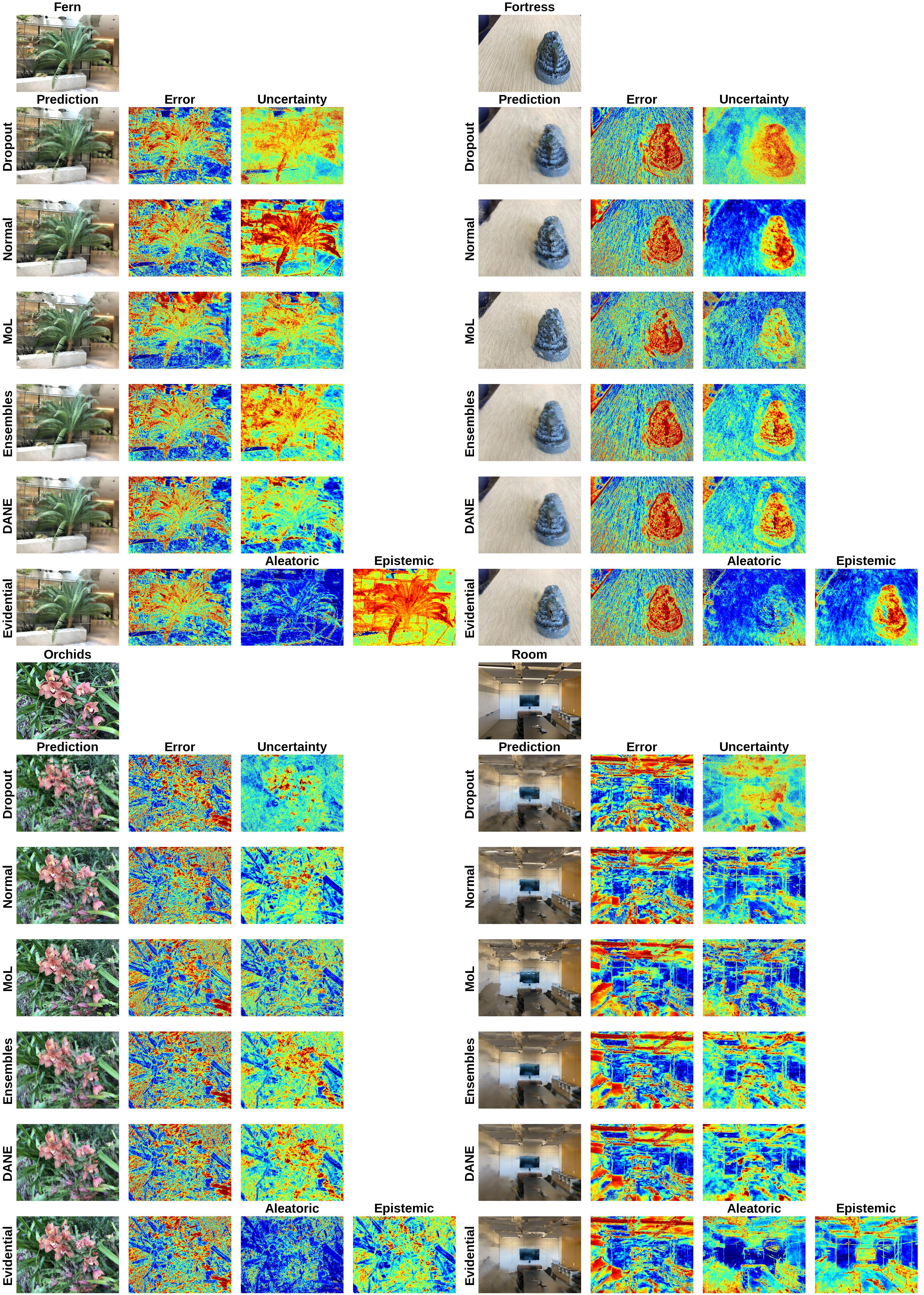}
    \caption{
        Qualitative comparison on LLFF.
    }
    \label{fig:qualitative_llff}
\end{figure*}

\begin{figure*}[t]
    \centering
    \includegraphics[height=\textheight]{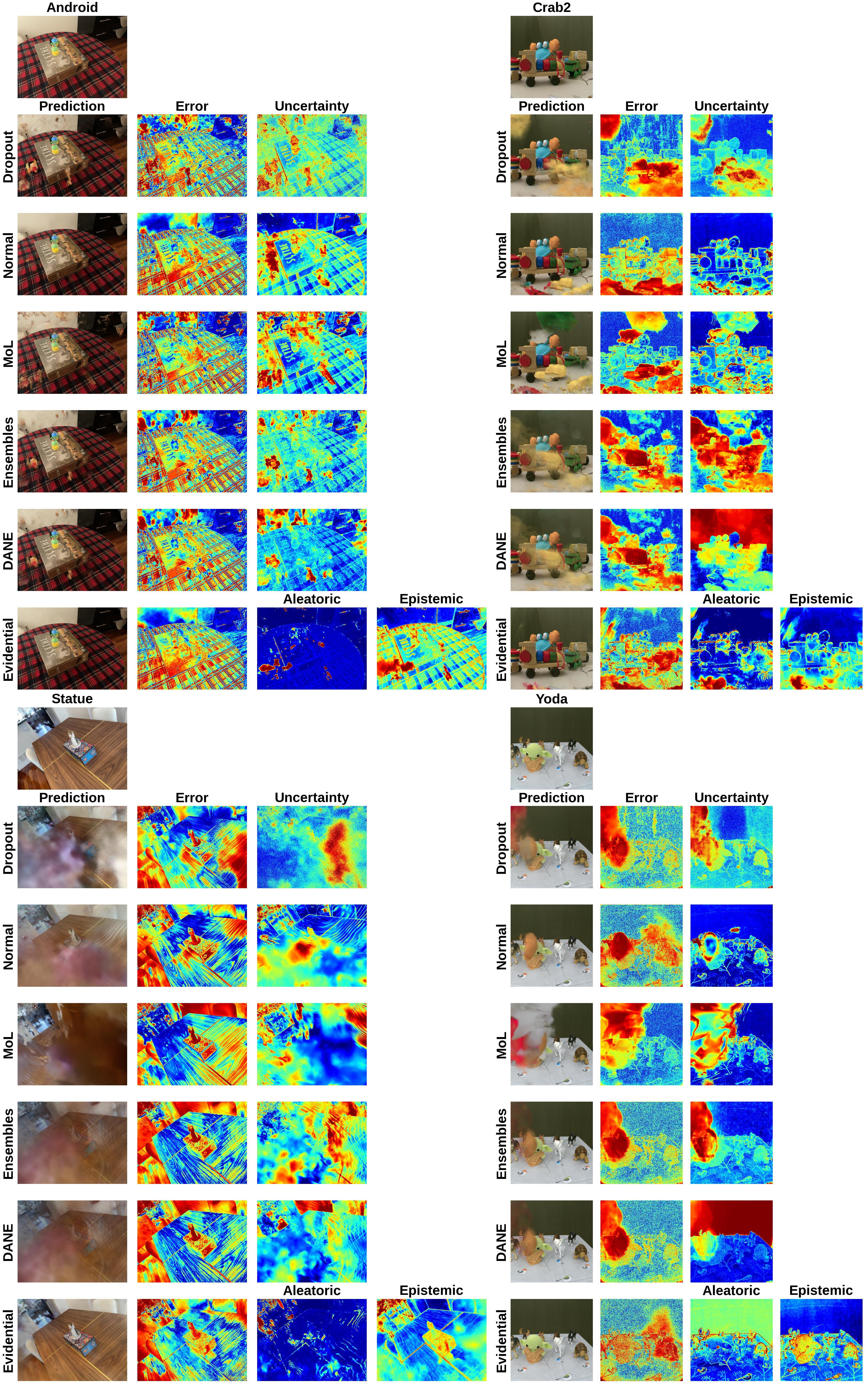}
    \caption{
        Qualitative comparison on RobustNeRF.
    }
    \label{fig:qualitative_robustnerf}
\end{figure*}

\begin{figure}[t]
    \centering
    \includegraphics[height=0.96\textheight, keepaspectratio]{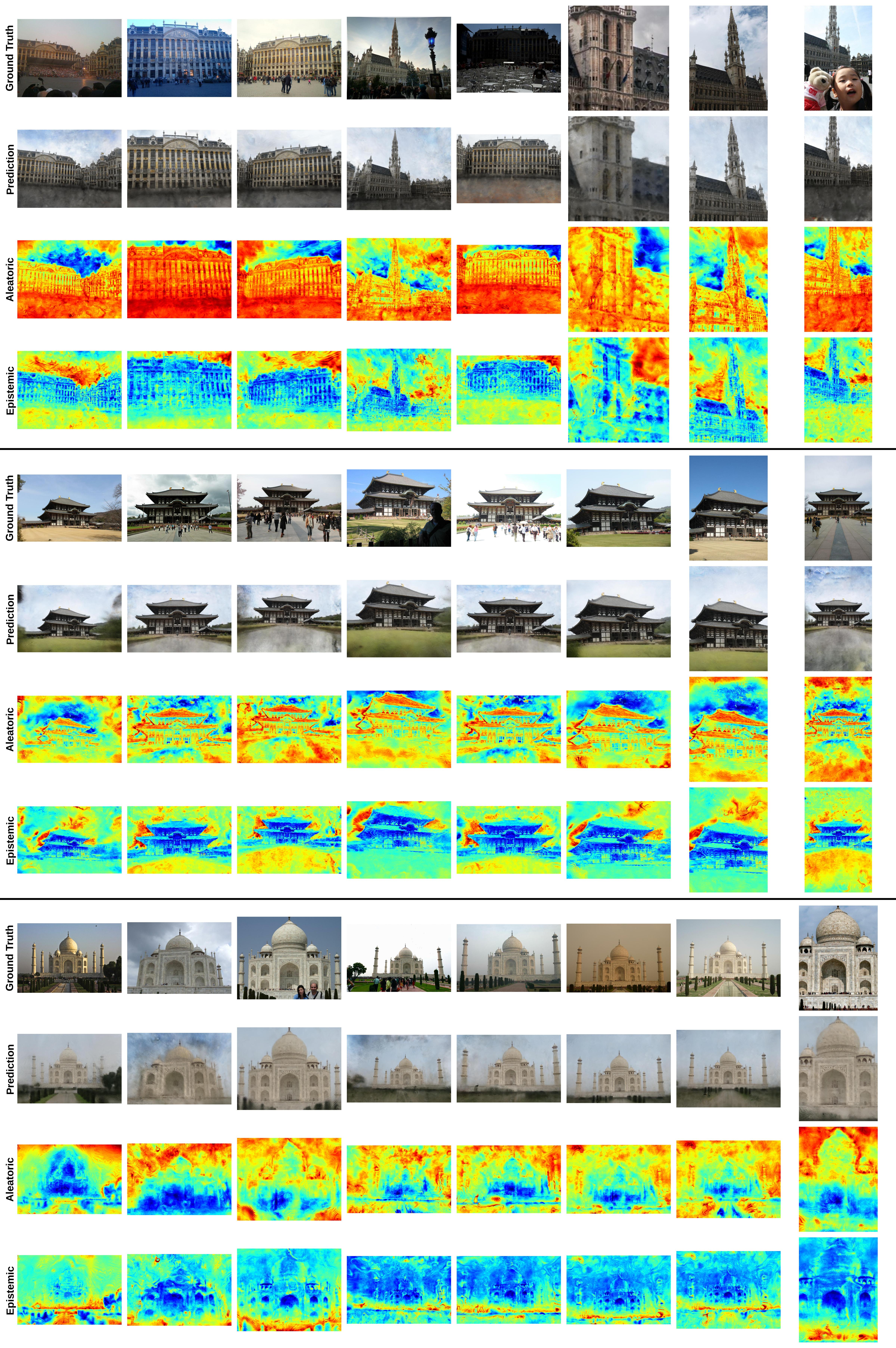}
    \caption{
        Aleatoric and epistemic uncertainty maps for three in-the-wild scenes from Phototourism.
        Aleatoric uncertainty peaks in regions with high radiance variance (e.g., sky higher above, building facades, moving figures), whereas epistemic uncertainty is concentrated in areas frequently occluded (e.g., sky directly behind the buildings or objects obscured by pedestrians).
    }
    \label{fig:qualitative_phototourism}
\end{figure}

\paragraphtitle{More qualitative results.}
We present the qualitative comparison of the uncertainty methods on LF, LLFF, and RobustNeRF datasets in~\Cref{fig:qualitative_lf,fig:qualitative_llff,fig:qualitative_robustnerf}, respectively.
Additionally, we show the aleatoric and epistemic uncertainty maps of more scenes in the wild from Phototoursim~\cite{phototourism} in~\Cref{fig:qualitative_phototourism}.

\begin{figure*}[h!]
    \centering
    \includegraphics[width=0.9\linewidth]{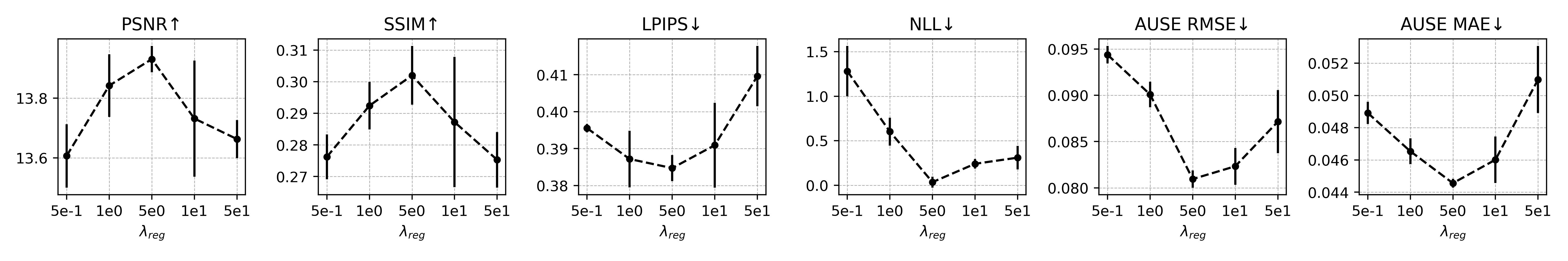}
    \vspace{-5mm}
    \caption{
        Sensitivity study of regularization coefficient's effect on all the quantitative metrics of \emph{Leaves} scene.
    }
    \label{fig:regularization}
\end{figure*}

\paragraphtitle{Hyperparameter selection.}
The regularization coefficient in the loss function is selected based on the quantitative metrics.
\Cref{fig:regularization} illustrates how different coefficients affect image reconstruction and uncertainty estimation.
In general, both excessively small and large coefficients lead to suboptimal performance, and the best hyperparameter value is inherently scene-dependent.
The specific regularization coefficients utilized to produce the reported results are detailed in \Cref{tab:regularization}.

\vspace{-4mm}

\vspace{-6mm}
\begin{table*}[h!]
    \centering
    \resizebox{\columnwidth}{!}{
        \newcolumntype{M}{>{\centering\arraybackslash}c{0.095\columnwidth}}
        \begin{tabular}{c | *{4}{c} | *{8}{c} | *{4}{c}}
        \toprule[1pt]
            Scene
            &Africa
            &Basket
            &Statue
            &Torch
            
            &Fern
            &Flower
            &Fortress
            &Horns
            &Leaves
            &Orchids
            &Room
            &T-Rex
            
            &Android
            &Crab
            &Statue
            &Yoda
            
        \\ \midrule
            $\lambda_\text{reg}$
            &5e-3
            &1e-1
            &1e-3
            &1e0

            &1e0
            &5e-2
            &1e0
            &5e-1
            &5e0
            &1e0
            &5e-1
            &5e0

            &5e-3
            &5e-4
            &1e-4
            &1e-5
        
        \\ \bottomrule[1pt]
        \end{tabular}
    }
    \vspace{-3mm}
    \caption{
        The regularization coefficients used in each scene of LF, LLFF, and RobustNeRF datasets.
    }
    \label{tab:regularization}
\end{table*}

\vspace{-4mm}

\paragraphtitle{Mutual causes of aleatoric and epistemic uncertainties.}
Various complex elements in a 3D scene can lead to elevated levels of either aleatoric or epistemic uncertainty.
However, attributing each specific factor exclusively to one type of uncertainty is often inappropriate, as many factors affect both uncertainties in different ways and to varying degrees.
For example, transient objects increase AU due to color variations introduced by motion, while simultaneously raising EU through partial occlusions.
Similarly, edges or high-frequency non-smooth regions tend to exhibit higher AU since their radiance is highly sensitive to input rays, as small inaccuracies in sensing, digitization, or poses can yield large radiance variations, resulting in nearly irreducible data uncertainty; meanwhile, the irregular geometry of such regions obstructs ray coverage and limits supervision signals from those surfaces, thereby increasing EU as well.

\begin{figure*}[h!]
    \centering
    \includegraphics[width=0.83\linewidth]{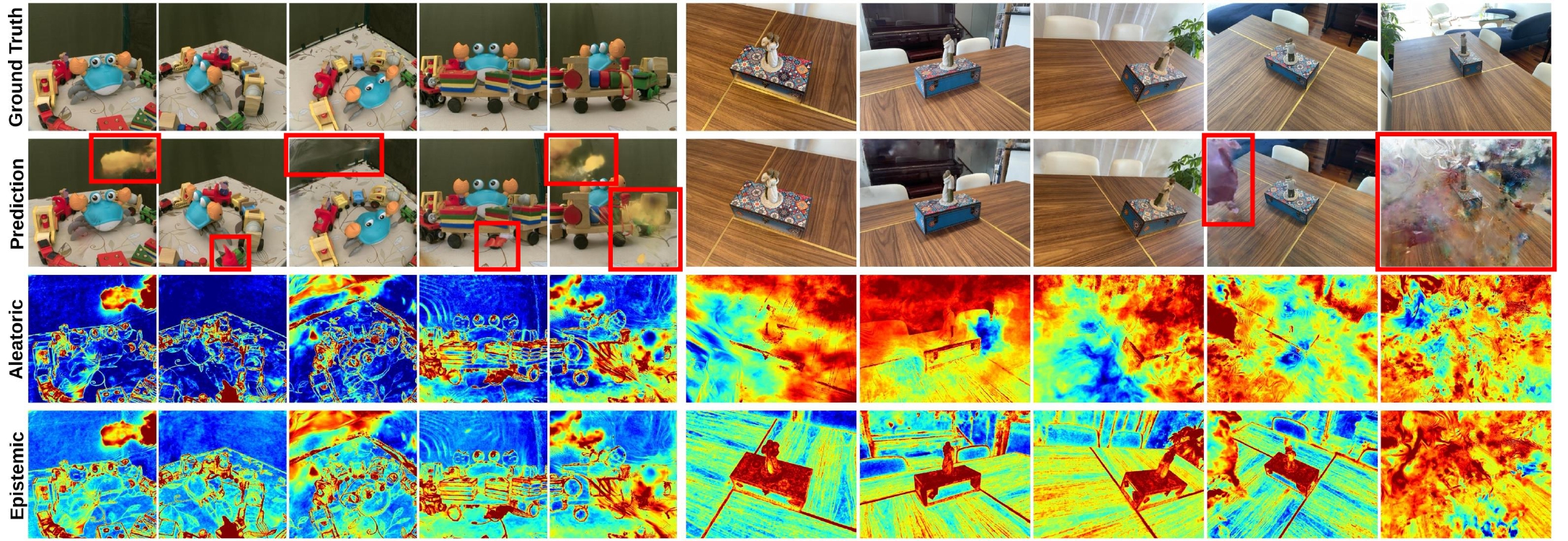}
    \vspace{-3mm}
    \caption{
        Aleatoric and epistemic uncertainties of scenes with transient objects.
        The red bounding boxes delineate the erroneous artifacts in test renderings caused by transients in the training views.
        If the model fails to suppress the floaters, both AU and EU are elevated on the transients;
        If the model resolves the transience, only AU is higher on the regions where the training transients were once present.
    }
    \label{fig:transients}
\end{figure*}

\paragraphtitle{Uncertainties of transients.}
Transient objects can lead to both higher aleatoric and epistemic uncertainties.
In practice, the uncertainties of the transient regions depend on the densities assigned to them by the model.
\Cref{fig:transients} shows ten images from two RobustNeRF scenes where the model is trained on images with cluttered objects.
It can be observed that when the model cannot disambiguate the transients and the floaters appear in the test renderings, both AU and EU tend to be higher on them, meaning that the model simultaneously receives inconsistent radiance signals (high AU) and lacks sufficient knowledge to determine the presence or geometry of the transients (high EU).
When the model resolves the transient objects (by minimizing their densities and removing them from volumetric rendering), the floaters disappear in the test image reconstructions and only AU remains high, indicating that the model no longer lacks the knowledge to determine the presence of the transients but still records the color inconsistency from the training signals as high AU.

\vspace{-3mm}

\clearpage
\begin{table*}[t]
    \newcommand{\meanstd}[2]{\shortstack{#1 \\ {\scriptsize \color{gray} $\pm$#2}}}
    \centering
    \resizebox{\textwidth}{!}{
        \newcolumntype{M}{>{\centering\arraybackslash}m{0.095\textwidth}}
        \begin{tabular}{cc|*{6}{M}}
        \toprule[1pt]
            Scene
            &Method
            &PSNR$\uparrow$
            &SSIM$\uparrow$
            &LPIPS$\downarrow$
            &NLL$\downarrow$
            &AUSE\newline RMSE$\downarrow$
            &AUSE\newline MAE$\downarrow$
        \\
        \midrule
            &Baseline
            &\meanstd{26.5406}{0.1903}
            &\meanstd{0.9019}{0.0009}
            &\meanstd{0.0565}{0.0010}
        \\
            &Dropout
            &\meanstd{26.6798}{0.0205}
            &\meanstd{0.8986}{0.0007}
            &\meanstd{0.0590}{0.0014}
            &\meanstd{4.0201}{0.2676}
            &\meanstd{0.0148}{0.0003}
            &\meanstd{0.0064}{0.0001}
        \\
            &Normal
            &\meanstd{27.4766}{1.5843}
            &\meanstd{0.9028}{0.0192}
            &\meanstd{0.0608}{0.0210}
            &\meanstd{-1.4911}{0.6587}
            &\meanstd{0.0078}{0.0023}
            &\meanstd{0.0033}{0.0004}
        \\
            Africa
            &MoL
            &\meanstd{28.0663}{0.6656}
            &\meanstd{0.9023}{0.0082}
            &\meanstd{0.0614}{0.0056}
            &\meanstd{-2.3375}{0.0088}
            &\meanstd{0.0069}{0.0001}
            &\meanstd{0.0035}{0.0001}
        \\
            &Ensembles
            &\meanstd{27.0083}{0.0652}
            &\meanstd{0.9171}{0.0003}
            &\meanstd{0.0491}{0.0006}
            &\meanstd{0.0871}{0.1899}
            &\meanstd{0.0087}{0.0005}
            &\meanstd{0.0037}{0.0001}
        \\
            &DANE
            &\meanstd{27.0083}{0.0652}
            &\meanstd{0.9171}{0.0003}
            &\meanstd{0.0491}{0.0006}
            &\meanstd{-0.9581}{0.1721}
            &\meanstd{0.0139}{0.0010}
            &\meanstd{0.0059}{0.0003}
        \\
            &Evidential
            &\meanstd{29.8826}{0.0617}
            &\meanstd{0.9331}{0.0002}
            &\meanstd{0.0340}{0.0015}
            &\meanstd{-2.3643}{0.0030}
            &\meanstd{0.0054}{0.0000}
            &\meanstd{0.0028}{0.0000}
        \\
        \midrule
            &Baseline
            &\meanstd{28.0171}{0.0703}
            &\meanstd{0.9114}{0.0019}
            &\meanstd{0.0474}{0.0008}
        \\
            &Dropout
            &\meanstd{27.3038}{0.0176}
            &\meanstd{0.8897}{0.0018}
            &\meanstd{0.0610}{0.0016}
            &\meanstd{6.2637}{0.3747}
            &\meanstd{0.0155}{0.0005}
            &\meanstd{0.0054}{0.0002}
        \\
            &Normal
            &\meanstd{27.9387}{0.6029}
            &\meanstd{0.9087}{0.0097}
            &\meanstd{0.0517}{0.0044}
            &\meanstd{5.5852}{2.7302}
            &\meanstd{0.0137}{0.0033}
            &\meanstd{0.0043}{0.0013}
        \\
            Basket
            &MoL
            &\meanstd{27.5847}{0.2379}
            &\meanstd{0.9044}{0.0007}
            &\meanstd{0.0692}{0.0015}
            &\meanstd{-2.2614}{0.0438}
            &\meanstd{0.0125}{0.0016}
            &\meanstd{0.0036}{0.0002}
        \\
            &Ensembles
            &\meanstd{28.9951}{0.0715}
            &\meanstd{0.9258}{0.0011}
            &\meanstd{0.0427}{0.0004}
            &\meanstd{-0.8529}{0.3663}
            &\meanstd{0.0058}{0.0002}
            &\meanstd{0.0023}{0.0001}
        \\
            &DANE
            &\meanstd{28.9951}{0.0715}
            &\meanstd{0.9258}{0.0011}
            &\meanstd{0.0427}{0.0004}
            &\meanstd{-1.0245}{0.3048}
            &\meanstd{0.0087}{0.0010}
            &\meanstd{0.0034}{0.0004}
        \\
            &Evidential
            &\meanstd{29.1442}{0.1841}
            &\meanstd{0.9263}{0.0008}
            &\meanstd{0.0369}{0.0002}
            &\meanstd{-2.1757}{0.0901}
            &\meanstd{0.0098}{0.0008}
            &\meanstd{0.0033}{0.0001}
        \\
        \midrule
            &Baseline
            &\meanstd{32.8018}{0.1611}
            &\meanstd{0.9645}{0.0007}
            &\meanstd{0.0221}{0.0007}
        \\
            &Dropout
            &\meanstd{31.8005}{0.0787}
            &\meanstd{0.9525}{0.0009}
            &\meanstd{0.0377}{0.0007}
            &\meanstd{-1.1505}{0.0358}
            &\meanstd{0.0057}{0.0002}
            &\meanstd{0.0031}{0.0001}
        \\
            &Normal
            &\meanstd{28.8110}{1.1896}
            &\meanstd{0.9457}{0.0036}
            &\meanstd{0.0424}{0.0033}
            &\meanstd{-1.9682}{0.4526}
            &\meanstd{0.0055}{0.0010}
            &\meanstd{0.0017}{0.0001}
        \\
            Statue
            &MoL
            &\meanstd{30.6262}{0.5385}
            &\meanstd{0.9469}{0.0083}
            &\meanstd{0.0545}{0.0174}
            &\meanstd{-2.9409}{0.0318}
            &\meanstd{0.0030}{0.0002}
            &\meanstd{0.0016}{0.0001}
        \\
            &Ensembles
            &\meanstd{33.7663}{0.0662}
            &\meanstd{0.9718}{0.0001}
            &\meanstd{0.0198}{0.0000}
            &\meanstd{-1.9914}{0.0994}
            &\meanstd{0.0029}{0.0001}
            &\meanstd{0.0017}{0.0000}
        \\
            &DANE
            &\meanstd{33.7663}{0.0662}
            &\meanstd{0.9718}{0.0001}
            &\meanstd{0.0198}{0.0000}
            &\meanstd{-2.1733}{0.0538}
            &\meanstd{0.0034}{0.0002}
            &\meanstd{0.0020}{0.0001}
        \\
            &Evidential
            &\meanstd{32.8506}{0.2748}
            &\meanstd{0.9661}{0.0003}
            &\meanstd{0.0221}{0.0003}
            &\meanstd{-2.9793}{0.0192}
            &\meanstd{0.0031}{0.0001}
            &\meanstd{0.0014}{0.0000}
        \\
        \midrule
            &Baseline
            &\meanstd{26.8557}{0.0802}
            &\meanstd{0.8911}{0.0010}
            &\meanstd{0.0599}{0.0014}
        \\
            &Dropout
            &\meanstd{26.6709}{0.0290}
            &\meanstd{0.8835}{0.0005}
            &\meanstd{0.0698}{0.0007}
            &\meanstd{5.5346}{0.4652}
            &\meanstd{0.0139}{0.0005}
            &\meanstd{0.0046}{0.0001}
        \\
            &Normal
            &\meanstd{27.7991}{0.1333}
            &\meanstd{0.9087}{0.0019}
            &\meanstd{0.0574}{0.0015}
            &\meanstd{-0.3559}{1.1275}
            &\meanstd{0.0090}{0.0022}
            &\meanstd{0.0023}{0.0002}
        \\
            Torch
            &MoL
            &\meanstd{26.6030}{0.1090}
            &\meanstd{0.8845}{0.0028}
            &\meanstd{0.0838}{0.0042}
            &\meanstd{-2.6173}{0.0145}
            &\meanstd{0.0112}{0.0005}
            &\meanstd{0.0028}{0.0000}
        \\
            &Ensembles
            &\meanstd{27.7419}{0.0417}
            &\meanstd{0.9086}{0.0004}
            &\meanstd{0.0528}{0.0002}
            &\meanstd{4.0553}{1.4211}
            &\meanstd{0.0105}{0.0006}
            &\meanstd{0.0026}{0.0001}
        \\
            &DANE
            &\meanstd{27.7419}{0.0417}
            &\meanstd{0.9086}{0.0004}
            &\meanstd{0.0528}{0.0002}
            &\meanstd{2.4292}{0.9753}
            &\meanstd{0.0144}{0.0013}
            &\meanstd{0.0044}{0.0005}
        \\
            &Evidential
            &\meanstd{27.9943}{0.1674}
            &\meanstd{0.9126}{0.0017}
            &\meanstd{0.0507}{0.0041}
            &\meanstd{-2.2771}{0.1350}
            &\meanstd{0.0096}{0.0022}
            &\meanstd{0.0025}{0.0003}
        \\
        \bottomrule[1pt]
        \end{tabular}
    }
    \caption{
        Mean and standard deviation of quantitative metrics over three runs on LF.
    }
    \label{tab:quantitative_lf}
\end{table*}
\begin{table*}[t]
    \newcommand{\meanstd}[2]{\shortstack{#1 \\ {\scriptsize \color{gray} $\pm$#2}}}
    \centering
    \resizebox{\textwidth}{!}{
        \newcolumntype{M}{>{\centering\arraybackslash}m{0.095\textwidth}}
        \begin{tabular}{cc|*{6}{M}}
        \toprule[1pt]
            Scene
            &Method
            &PSNR$\uparrow$
            &SSIM$\uparrow$
            &LPIPS$\downarrow$
            &NLL$\downarrow$
            &AUSE\newline RMSE$\downarrow$
            &AUSE\newline MAE$\downarrow$
        \\
        \midrule
            &Baseline
            &\meanstd{20.1069}{0.1190}
            &\meanstd{0.5685}{0.0005}
            &\meanstd{0.3617}{0.0050}
        \\
            &Dropout
            &\meanstd{19.1317}{0.1808}
            &\meanstd{0.5153}{0.0059}
            &\meanstd{0.4432}{0.0060}
            &\meanstd{46.5072}{8.3738}
            &\meanstd{0.0517}{0.0016}
            &\meanstd{0.0294}{0.0020}
        \\
            &Normal
            &\meanstd{20.1763}{0.3308}
            &\meanstd{0.5905}{0.0115}
            &\meanstd{0.3273}{0.0133}
            &\meanstd{47.5151}{1.6352}
            &\meanstd{0.0349}{0.0014}
            &\meanstd{0.0193}{0.0005}
        \\
            Fern
            &MoL
            &\meanstd{18.2953}{0.6532}
            &\meanstd{0.5222}{0.0298}
            &\meanstd{0.4279}{0.0406}
            &\meanstd{1.3336}{0.2902}
            &\meanstd{0.0605}{0.0157}
            &\meanstd{0.0257}{0.0078}
        \\
            &Ensembles
            &\meanstd{20.7328}{0.0591}
            &\meanstd{0.6134}{0.0016}
            &\meanstd{0.3462}{0.0019}
            &\meanstd{3.2024}{0.4579}
            &\meanstd{0.0297}{0.0004}
            &\meanstd{0.0148}{0.0002}
        \\
            &DANE
            &\meanstd{20.7328}{0.0591}
            &\meanstd{0.6134}{0.0016}
            &\meanstd{0.3462}{0.0019}
            &\meanstd{2.6334}{0.4099}
            &\meanstd{0.0332}{0.0002}
            &\meanstd{0.0172}{0.0001}
        \\
            &Evidential
            &\meanstd{20.8095}{0.1119}
            &\meanstd{0.6216}{0.0074}
            &\meanstd{0.3005}{0.0077}
            &\meanstd{-0.4856}{0.1133}
            &\meanstd{0.0301}{0.0009}
            &\meanstd{0.0160}{0.0007}
        \\
        \midrule
            &Baseline
            &\meanstd{18.3456}{0.0046}
            &\meanstd{0.4792}{0.0068}
            &\meanstd{0.4230}{0.0116}
        \\
            &Dropout
            &\meanstd{18.6701}{0.1790}
            &\meanstd{0.4990}{0.0089}
            &\meanstd{0.3780}{0.0057}
            &\meanstd{54.1338}{2.6904}
            &\meanstd{0.0554}{0.0012}
            &\meanstd{0.0341}{0.0012}
        \\
            &Normal
            &\meanstd{12.6188}{4.7635}
            &\meanstd{0.2205}{0.1075}
            &\meanstd{0.7529}{0.2290}
            &\meanstd{9.0113}{6.3991}
            &\meanstd{0.0732}{0.0073}
            &\meanstd{0.0590}{0.0115}
        \\
            Flower
            &MoL
            &\meanstd{18.6338}{0.2226}
            &\meanstd{0.5140}{0.0055}
            &\meanstd{0.3665}{0.0141}
            &\meanstd{1.4129}{0.1601}
            &\meanstd{0.0589}{0.0023}
            &\meanstd{0.0296}{0.0012}
        \\
            &Ensembles
            &\meanstd{18.7711}{0.0344}
            &\meanstd{0.5204}{0.0036}
            &\meanstd{0.3993}{0.0050}
            &\meanstd{7.7562}{0.2043}
            &\meanstd{0.0370}{0.0004}
            &\meanstd{0.0173}{0.0002}
        \\
            &DANE
            &\meanstd{18.7711}{0.0344}
            &\meanstd{0.5204}{0.0036}
            &\meanstd{0.3993}{0.0050}
            &\meanstd{7.7131}{0.2053}
            &\meanstd{0.0371}{0.0004}
            &\meanstd{0.0174}{0.0002}
        \\
            &Evidential
            &\meanstd{19.2606}{0.1150}
            &\meanstd{0.5254}{0.0126}
            &\meanstd{0.3957}{0.0208}
            &\meanstd{2.4728}{0.6069}
            &\meanstd{0.0409}{0.0022}
            &\meanstd{0.0238}{0.0015}
        \\
        \midrule
            &Baseline
            &\meanstd{18.3478}{0.1598}
            &\meanstd{0.3906}{0.0033}
            &\meanstd{0.5240}{0.0069}
        \\
            &Dropout
            &\meanstd{18.3708}{0.5357}
            &\meanstd{0.4063}{0.0039}
            &\meanstd{0.5204}{0.0077}
            &\meanstd{52.6142}{9.0302}
            &\meanstd{0.0518}{0.0069}
            &\meanstd{0.0278}{0.0022}
        \\
            &Normal
            &\meanstd{18.3146}{0.0871}
            &\meanstd{0.4100}{0.0106}
            &\meanstd{0.5060}{0.0427}
            &\meanstd{62.8832}{20.7391}
            &\meanstd{0.0515}{0.0036}
            &\meanstd{0.0248}{0.0027}
        \\
            Fortress
            &MoL
            &\meanstd{17.4740}{0.1443}
            &\meanstd{0.3779}{0.0064}
            &\meanstd{0.6090}{0.0253}
            &\meanstd{1.8754}{0.5340}
            &\meanstd{0.0645}{0.0009}
            &\meanstd{0.0280}{0.0003}
        \\
            &Ensembles
            &\meanstd{18.8273}{0.0496}
            &\meanstd{0.4438}{0.0015}
            &\meanstd{0.4517}{0.0023}
            &\meanstd{7.5589}{0.8103}
            &\meanstd{0.0369}{0.0014}
            &\meanstd{0.0188}{0.0003}
        \\
            &DANE
            &\meanstd{18.8273}{0.0496}
            &\meanstd{0.4438}{0.0015}
            &\meanstd{0.4517}{0.0023}
            &\meanstd{7.1596}{0.8387}
            &\meanstd{0.0377}{0.0017}
            &\meanstd{0.0195}{0.0003}
        \\
            &Evidential
            &\meanstd{18.7310}{0.0406}
            &\meanstd{0.4264}{0.0065}
            &\meanstd{0.4878}{0.0040}
            &\meanstd{0.2943}{0.1765}
            &\meanstd{0.0404}{0.0011}
            &\meanstd{0.0208}{0.0005}
        \\
        \midrule
            &Baseline
            &\meanstd{15.7052}{0.1371}
            &\meanstd{0.4527}{0.0105}
            &\meanstd{0.4451}{0.0111}
        \\
            &Dropout
            &\meanstd{15.4649}{0.2658}
            &\meanstd{0.4296}{0.0113}
            &\meanstd{0.4729}{0.0074}
            &\meanstd{159.8088}{22.3326}
            &\meanstd{0.0919}{0.0057}
            &\meanstd{0.0482}{0.0025}
        \\
            &Normal
            &\meanstd{13.4923}{0.2916}
            &\meanstd{0.2336}{0.0089}
            &\meanstd{0.6160}{0.0096}
            &\meanstd{22.6808}{7.0886}
            &\meanstd{0.1130}{0.0047}
            &\meanstd{0.0807}{0.0083}
        \\
            Horns
            &MoL
            &\meanstd{14.5633}{0.1664}
            &\meanstd{0.3638}{0.0457}
            &\meanstd{0.5691}{0.0891}
            &\meanstd{2.1617}{0.2982}
            &\meanstd{0.1101}{0.0039}
            &\meanstd{0.0505}{0.0008}
        \\
            &Ensembles
            &\meanstd{15.9843}{0.0172}
            &\meanstd{0.5025}{0.0033}
            &\meanstd{0.4240}{0.0045}
            &\meanstd{22.0785}{1.0690}
            &\meanstd{0.0759}{0.0013}
            &\meanstd{0.0328}{0.0005}
        \\
            &DANE
            &\meanstd{15.9843}{0.0172}
            &\meanstd{0.5025}{0.0033}
            &\meanstd{0.4240}{0.0045}
            &\meanstd{15.3955}{1.4795}
            &\meanstd{0.0750}{0.0014}
            &\meanstd{0.0330}{0.0006}
        \\
            &Evidential
            &\meanstd{15.7596}{0.1745}
            &\meanstd{0.5034}{0.0074}
            &\meanstd{0.3878}{0.0075}
            &\meanstd{1.8495}{0.1097}
            &\meanstd{0.0941}{0.0055}
            &\meanstd{0.0403}{0.0023}
        \\
        \bottomrule[1pt]
        \end{tabular}
    }
    \caption{
        Mean and standard deviation of quantitative metrics over three runs on LLFF.
    }
    \label{tab:quantitative_llff1}
\end{table*}
\begin{table*}[t]
    \newcommand{\meanstd}[2]{\shortstack{#1 \\ {\scriptsize \color{gray} $\pm$#2}}}
    \centering
    \resizebox{\textwidth}{!}{
        \newcolumntype{M}{>{\centering\arraybackslash}m{0.095\textwidth}}
        \begin{tabular}{cc|*{6}{M}}
        \toprule[1pt]
            Scene
            &Method
            &PSNR$\uparrow$
            &SSIM$\uparrow$
            &LPIPS$\downarrow$
            &NLL$\downarrow$
            &AUSE\newline RMSE$\downarrow$
            &AUSE\newline MAE$\downarrow$
        \\
        \midrule
            &Baseline
            &\meanstd{13.6976}{0.1464}
            &\meanstd{0.2557}{0.0155}
            &\meanstd{0.4167}{0.0066}
        \\
            &Dropout
            &\meanstd{13.9549}{0.2228}
            &\meanstd{0.2423}{0.0248}
            &\meanstd{0.4235}{0.0211}
            &\meanstd{141.2669}{12.2890}
            &\meanstd{0.1072}{0.0025}
            &\meanstd{0.0618}{0.0024}
        \\
            &Normal
            &\meanstd{13.3743}{0.0707}
            &\meanstd{0.2525}{0.0111}
            &\meanstd{0.4239}{0.0075}
            &\meanstd{104.3151}{41.1687}
            &\meanstd{0.1064}{0.0013}
            &\meanstd{0.0553}{0.0009}
        \\
            Leaves
            &MoL
            &\meanstd{12.8939}{0.1964}
            &\meanstd{0.2408}{0.0179}
            &\meanstd{0.4276}{0.0162}
            &\meanstd{4.1014}{0.2543}
            &\meanstd{0.1266}{0.0053}
            &\meanstd{0.0674}{0.0044}
        \\
            &Ensembles
            &\meanstd{14.2397}{0.0569}
            &\meanstd{0.2960}{0.0069}
            &\meanstd{0.4319}{0.0047}
            &\meanstd{12.3460}{0.5912}
            &\meanstd{0.0915}{0.0010}
            &\meanstd{0.0471}{0.0007}
        \\
            &DANE
            &\meanstd{14.2397}{0.0569}
            &\meanstd{0.2960}{0.0069}
            &\meanstd{0.4319}{0.0047}
            &\meanstd{11.6017}{0.1579}
            &\meanstd{0.0928}{0.0009}
            &\meanstd{0.0482}{0.0006}
        \\
            &Evidential
            &\meanstd{13.9301}{0.0437}
            &\meanstd{0.3020}{0.0093}
            &\meanstd{0.3847}{0.0035}
            &\meanstd{0.0349}{0.0607}
            &\meanstd{0.0809}{0.0009}
            &\meanstd{0.0446}{0.0003}
        \\
        \midrule
            &Baseline
            &\meanstd{14.5292}{0.0830}
            &\meanstd{0.3036}{0.0057}
            &\meanstd{0.3916}{0.0103}
        \\
            &Dropout
            &\meanstd{13.3110}{0.4149}
            &\meanstd{0.2151}{0.0202}
            &\meanstd{0.4709}{0.0100}
            &\meanstd{123.9834}{10.5705}
            &\meanstd{0.1195}{0.0056}
            &\meanstd{0.0762}{0.0037}
        \\
            &Normal
            &\meanstd{14.9466}{0.1068}
            &\meanstd{0.3189}{0.0126}
            &\meanstd{0.3911}{0.0160}
            &\meanstd{25.0742}{5.0559}
            &\meanstd{0.0754}{0.0028}
            &\meanstd{0.0424}{0.0013}
        \\
            Orchids
            &MoL
            &\meanstd{13.8818}{0.0307}
            &\meanstd{0.3242}{0.0031}
            &\meanstd{0.3706}{0.0047}
            &\meanstd{2.4593}{0.0292}
            &\meanstd{0.1058}{0.0045}
            &\meanstd{0.0506}{0.0021}
        \\
            &Ensembles
            &\meanstd{14.8488}{0.1379}
            &\meanstd{0.3250}{0.0081}
            &\meanstd{0.3953}{0.0065}
            &\meanstd{9.7556}{0.8979}
            &\meanstd{0.0753}{0.0037}
            &\meanstd{0.0409}{0.0017}
        \\
            &DANE
            &\meanstd{14.8488}{0.1379}
            &\meanstd{0.3250}{0.0081}
            &\meanstd{0.3953}{0.0065}
            &\meanstd{9.4356}{0.7865}
            &\meanstd{0.0746}{0.0034}
            &\meanstd{0.0407}{0.0016}
        \\
            &Evidential
            &\meanstd{14.7012}{0.1548}
            &\meanstd{0.3253}{0.0090}
            &\meanstd{0.3601}{0.0103}
            &\meanstd{0.5681}{0.2424}
            &\meanstd{0.0843}{0.0041}
            &\meanstd{0.0459}{0.0017}
        \\
        \midrule
            &Baseline
            &\meanstd{19.7859}{0.0325}
            &\meanstd{0.7196}{0.0037}
            &\meanstd{0.3864}{0.0088}
        \\
            &Dropout
            &\meanstd{19.1522}{0.0418}
            &\meanstd{0.6704}{0.0097}
            &\meanstd{0.4536}{0.0085}
            &\meanstd{79.6637}{8.4028}
            &\meanstd{0.0637}{0.0016}
            &\meanstd{0.0342}{0.0006}
        \\
            &Normal
            &\meanstd{19.8485}{0.2959}
            &\meanstd{0.6766}{0.0384}
            &\meanstd{0.4178}{0.0630}
            &\meanstd{111.9164}{13.8320}
            &\meanstd{0.0465}{0.0013}
            &\meanstd{0.0232}{0.0020}
        \\
            Room
            &MoL
            &\meanstd{17.6348}{0.7025}
            &\meanstd{0.5337}{0.0754}
            &\meanstd{0.6094}{0.0874}
            &\meanstd{3.1864}{0.4821}
            &\meanstd{0.0651}{0.0042}
            &\meanstd{0.0295}{0.0026}
        \\
            &Ensembles
            &\meanstd{19.9259}{0.0347}
            &\meanstd{0.7492}{0.0003}
            &\meanstd{0.3656}{0.0006}
            &\meanstd{15.8667}{0.7134}
            &\meanstd{0.0347}{0.0006}
            &\meanstd{0.0161}{0.0004}
        \\
            &DANE
            &\meanstd{19.9259}{0.0347}
            &\meanstd{0.7492}{0.0003}
            &\meanstd{0.3656}{0.0006}
            &\meanstd{13.4646}{0.4077}
            &\meanstd{0.0368}{0.0004}
            &\meanstd{0.0174}{0.0005}
        \\
            &Evidential
            &\meanstd{19.9548}{0.0570}
            &\meanstd{0.7172}{0.0181}
            &\meanstd{0.3660}{0.0299}
            &\meanstd{1.2947}{1.1068}
            &\meanstd{0.0465}{0.0027}
            &\meanstd{0.0221}{0.0012}
        \\
        \midrule
            &Baseline
            &\meanstd{19.7067}{0.0095}
            &\meanstd{0.6063}{0.0041}
            &\meanstd{0.3463}{0.0047}
        \\
            &Dropout
            &\meanstd{19.0824}{0.1470}
            &\meanstd{0.5750}{0.0059}
            &\meanstd{0.4119}{0.0046}
            &\meanstd{67.4407}{5.4653}
            &\meanstd{0.0525}{0.0006}
            &\meanstd{0.0290}{0.0006}
        \\
            &Normal
            &\meanstd{19.6283}{0.1296}
            &\meanstd{0.6055}{0.0176}
            &\meanstd{0.3502}{0.0238}
            &\meanstd{59.4677}{29.4200}
            &\meanstd{0.0379}{0.0026}
            &\meanstd{0.0189}{0.0008}
        \\
            T-Rex
            &MoL
            &\meanstd{17.9974}{0.5803}
            &\meanstd{0.5127}{0.0837}
            &\meanstd{0.4935}{0.1085}
            &\meanstd{1.4452}{0.3356}
            &\meanstd{0.0594}{0.0049}
            &\meanstd{0.0255}{0.0029}
        \\
            &Ensembles
            &\meanstd{20.0150}{0.0311}
            &\meanstd{0.6373}{0.0019}
            &\meanstd{0.3317}{0.0028}
            &\meanstd{10.7621}{0.1102}
            &\meanstd{0.0293}{0.0004}
            &\meanstd{0.0145}{0.0002}
        \\
            &DANE
            &\meanstd{20.0150}{0.0311}
            &\meanstd{0.6373}{0.0019}
            &\meanstd{0.3317}{0.0028}
            &\meanstd{10.4152}{0.0855}
            &\meanstd{0.0295}{0.0004}
            &\meanstd{0.0146}{0.0002}
        \\
            &Evidential
            &\meanstd{19.8874}{0.0246}
            &\meanstd{0.6329}{0.0048}
            &\meanstd{0.3177}{0.0095}
            &\meanstd{-0.6168}{0.0407}
            &\meanstd{0.0455}{0.0004}
            &\meanstd{0.0226}{0.0006}
        \\
        \bottomrule[1pt]
        \end{tabular}
    }
    \caption{
        Mean and standard deviation of quantitative metrics over three runs on LLFF.
    }
    \label{tab:quantitative_llff2}
\end{table*}
\begin{table*}[t]
    \newcommand{\meanstd}[2]{\shortstack{#1 \\ {\scriptsize \color{gray} $\pm$#2}}}
    \centering
    \resizebox{\textwidth}{!}{
        \newcolumntype{M}{>{\centering\arraybackslash}m{0.095\textwidth}}
        \begin{tabular}{cc|*{6}{M}}
        \toprule[1pt]
            Scene
            &Method
            &PSNR$\uparrow$
            &SSIM$\uparrow$
            &LPIPS$\downarrow$
            &NLL$\downarrow$
            &AUSE\newline RMSE$\downarrow$
            &AUSE\newline MAE$\downarrow$
        \\
        \midrule
            &Baseline
            &\meanstd{22.8865}{0.0476}
            &\meanstd{0.7691}{0.0008}
            &\meanstd{0.1545}{0.0011}
        \\
            &Dropout
            &\meanstd{22.6094}{0.0779}
            &\meanstd{0.7552}{0.0001}
            &\meanstd{0.1699}{0.0009}
            &\meanstd{17.8212}{0.1534}
            &\meanstd{0.0308}{0.0002}
            &\meanstd{0.0188}{0.0001}
        \\
            &Normal
            &\meanstd{23.9819}{0.0770}
            &\meanstd{0.8146}{0.0013}
            &\meanstd{0.1058}{0.0021}
            &\meanstd{4.9525}{0.7433}
            &\meanstd{0.0227}{0.0006}
            &\meanstd{0.0152}{0.0003}
        \\
            Android
            &MoL
            &\meanstd{22.1825}{0.0186}
            &\meanstd{0.7395}{0.0002}
            &\meanstd{0.1905}{0.0008}
            &\meanstd{-0.7117}{0.0656}
            &\meanstd{0.0273}{0.0006}
            &\meanstd{0.0172}{0.0005}
        \\
            &Ensembles
            &\meanstd{23.4866}{0.0304}
            &\meanstd{0.8012}{0.0006}
            &\meanstd{0.1381}{0.0006}
            &\meanstd{8.1690}{0.8485}
            &\meanstd{0.0265}{0.0001}
            &\meanstd{0.0158}{0.0000}
        \\
            &DANE
            &\meanstd{23.4866}{0.0304}
            &\meanstd{0.8012}{0.0006}
            &\meanstd{0.1381}{0.0006}
            &\meanstd{7.4801}{0.8655}
            &\meanstd{0.0274}{0.0001}
            &\meanstd{0.0164}{0.0000}
        \\
            &Evidential
            &\meanstd{23.8915}{0.0651}
            &\meanstd{0.8116}{0.0034}
            &\meanstd{0.1047}{0.0007}
            &\meanstd{-1.1616}{0.0266}
            &\meanstd{0.0231}{0.0003}
            &\meanstd{0.0153}{0.0002}
        \\
        \midrule
            &Baseline
            &\meanstd{28.6708}{0.2399}
            &\meanstd{0.9034}{0.0038}
            &\meanstd{0.1124}{0.0072}
        \\
            &Dropout
            &\meanstd{27.5569}{0.3792}
            &\meanstd{0.8850}{0.0049}
            &\meanstd{0.1186}{0.0065}
            &\meanstd{24.1365}{2.5685}
            &\meanstd{0.0216}{0.0012}
            &\meanstd{0.0074}{0.0007}
        \\
            &Normal
            &\meanstd{27.7154}{0.2508}
            &\meanstd{0.9062}{0.0016}
            &\meanstd{0.0971}{0.0025}
            &\meanstd{2.9771}{1.6485}
            &\meanstd{0.0151}{0.0007}
            &\meanstd{0.0062}{0.0004}
        \\
            Crab
            &MoL
            &\meanstd{27.0696}{0.2685}
            &\meanstd{0.8767}{0.0044}
            &\meanstd{0.1374}{0.0061}
            &\meanstd{-2.5692}{0.0274}
            &\meanstd{0.0186}{0.0013}
            &\meanstd{0.0065}{0.0005}
        \\
            &Ensembles
            &\meanstd{30.0907}{0.0696}
            &\meanstd{0.9206}{0.0007}
            &\meanstd{0.1072}{0.0006}
            &\meanstd{-0.3212}{0.4515}
            &\meanstd{0.0048}{0.0005}
            &\meanstd{0.0021}{0.0001}
        \\
            &DANE
            &\meanstd{30.0907}{0.0696}
            &\meanstd{0.9206}{0.0007}
            &\meanstd{0.1072}{0.0006}
            &\meanstd{-0.0123}{0.2259}
            &\meanstd{0.0234}{0.0006}
            &\meanstd{0.0101}{0.0000}
        \\
            &Evidential
            &\meanstd{29.9324}{0.0145}
            &\meanstd{0.9160}{0.0086}
            &\meanstd{0.0795}{0.0154}
            &\meanstd{-2.0303}{0.2399}
            &\meanstd{0.0150}{0.0059}
            &\meanstd{0.0084}{0.0058}
        \\
        \midrule
            &Baseline
            &\meanstd{20.1787}{0.0379}
            &\meanstd{0.7489}{0.0046}
            &\meanstd{0.2356}{0.0072}
        \\
            &Dropout
            &\meanstd{20.0340}{0.0841}
            &\meanstd{0.7213}{0.0068}
            &\meanstd{0.2772}{0.0159}
            &\meanstd{35.4746}{2.1971}
            &\meanstd{0.0464}{0.0013}
            &\meanstd{0.0304}{0.0008}
        \\
            &Normal
            &\meanstd{20.1453}{0.0154}
            &\meanstd{0.7780}{0.0039}
            &\meanstd{0.2128}{0.0061}
            &\meanstd{17.9184}{2.0826}
            &\meanstd{0.0413}{0.0008}
            &\meanstd{0.0275}{0.0007}
        \\
            Statue
            &MoL
            &\meanstd{19.3304}{0.1637}
            &\meanstd{0.6834}{0.0032}
            &\meanstd{0.3300}{0.0069}
            &\meanstd{0.0773}{0.0274}
            &\meanstd{0.0571}{0.0005}
            &\meanstd{0.0380}{0.0007}
        \\
            &Ensembles
            &\meanstd{20.5674}{0.0059}
            &\meanstd{0.7843}{0.0008}
            &\meanstd{0.2166}{0.0028}
            &\meanstd{9.9047}{0.3519}
            &\meanstd{0.0297}{0.0003}
            &\meanstd{0.0191}{0.0001}
        \\
            &DANE
            &\meanstd{20.5674}{0.0059}
            &\meanstd{0.7843}{0.0008}
            &\meanstd{0.2166}{0.0028}
            &\meanstd{8.4305}{0.2763}
            &\meanstd{0.0367}{0.0002}
            &\meanstd{0.0235}{0.0000}
        \\
            &Evidential
            &\meanstd{20.8284}{0.0496}
            &\meanstd{0.8017}{0.0010}
            &\meanstd{0.1780}{0.0019}
            &\meanstd{-0.0313}{0.3149}
            &\meanstd{0.0362}{0.0008}
            &\meanstd{0.0243}{0.0004}
        \\
        \midrule
            &Baseline
            &\meanstd{29.1458}{0.4079}
            &\meanstd{0.8969}{0.0081}
            &\meanstd{0.1284}{0.0118}
        \\
            &Dropout
            &\meanstd{28.8770}{0.1990}
            &\meanstd{0.8857}{0.0041}
            &\meanstd{0.1322}{0.0020}
            &\meanstd{14.0873}{5.1782}
            &\meanstd{0.0148}{0.0034}
            &\meanstd{0.0073}{0.0014}
        \\
            &Normal
            &\meanstd{29.3523}{0.5839}
            &\meanstd{0.9099}{0.0040}
            &\meanstd{0.1088}{0.0043}
            &\meanstd{16.1012}{8.0429}
            &\meanstd{0.0209}{0.0034}
            &\meanstd{0.0115}{0.0018}
        \\
            Yoda
            &MoL
            &\meanstd{26.5671}{0.3613}
            &\meanstd{0.8688}{0.0051}
            &\meanstd{0.1617}{0.0070}
            &\meanstd{-2.3751}{0.0676}
            &\meanstd{0.0186}{0.0018}
            &\meanstd{0.0088}{0.0010}
        \\
            &Ensembles
            &\meanstd{30.6364}{0.1516}
            &\meanstd{0.9188}{0.0025}
            &\meanstd{0.1135}{0.0029}
            &\meanstd{0.7713}{0.1332}
            &\meanstd{0.0044}{0.0001}
            &\meanstd{0.0020}{0.0001}
        \\
            &DANE
            &\meanstd{30.6364}{0.1516}
            &\meanstd{0.9188}{0.0025}
            &\meanstd{0.1135}{0.0029}
            &\meanstd{0.5383}{0.2742}
            &\meanstd{0.0257}{0.0015}
            &\meanstd{0.0116}{0.0006}
        \\
            &Evidential
            &\meanstd{30.2646}{0.1857}
            &\meanstd{0.9272}{0.0065}
            &\meanstd{0.0826}{0.0124}
            &\meanstd{-1.8577}{0.2073}
            &\meanstd{0.0139}{0.0009}
            &\meanstd{0.0072}{0.0012}
        \\
        \bottomrule[1pt]
        \end{tabular}
    }
    \caption{
        Mean and standard deviation of quantitative metrics over three runs on RobustNeRF.
    }
    \label{tab:quantitative_robustnerf}
\end{table*}

\clearpage

\end{document}